\documentclass[journal]{IEEEtran}
\usepackage{amsmath,amsfonts}
\usepackage{algorithmic}
\usepackage{algorithm}
\usepackage{array}
\usepackage[caption=false,font=normalsize,labelfont=sf,textfont=sf]{subfig}
\usepackage{textcomp}
\usepackage{stfloats}
\usepackage{url}
\usepackage{verbatim}
\usepackage{graphicx}
\usepackage{authblk}

\usepackage{booktabs}

\usepackage[svgnames]{xcolor}

\usepackage[numbers,sort]{natbib}

\usepackage{pgfplots}
\pgfplotsset{compat=1.18}
\usetikzlibrary{positioning,arrows.meta,decorations.pathreplacing,calc}

\usepackage[hidelinks]{hyperref}

\begin{document}
\bstctlcite{IEEEexample:BSTcontrol}

\title{Multimodal Fusion with Pre-trained Models for Psychotherapy Outcome Prediction from Emotional Dynamics}

\author[1]{Tobias Hallmen\thanks{Corresponding author: Tobias Hallmen (tobias.hallmen@uni-a.de).}\thanks{This work has been submitted to the IEEE for possible publication. Copyright may be transferred without notice, after which this version may no longer be accessible.}}
\author[1]{Marius Funk}
\author[2]{Antonia Vehlen}
\author[2]{Jeanne Hierse}
\author[2]{Gerhard Prinz}
\author[2]{Steffen Eberhardt}
\author[2]{\\Wolfgang Lutz}
\author[1]{Elisabeth André}
\affil[1]{Chair for Human-Centered Artificial Intelligence, University of Augsburg}
\affil[2]{Clinical Psychology and Psychotherapy, University of Trier}

\markboth{IEEE Transactions on Affective Computing}{Hallmen \MakeLowercase{\textit{et al.}}: Multimodal Fusion with Pretrained Models for Psychotherapy Outcome Prediction}

\maketitle

\begin{abstract}
In-session emotional dynamics are associated with psychotherapy outcomes but difficult to measure at scale. Pretrained multimodal models provide automated access to facial, vocal, and textual signals, yet clinical datasets remain small. Our research question is how pretrained multimodal representations can best support psychotherapy outcome prediction under data scarcity. Three studies use 106 sessions from 88 patients and nine session-level outcome scales (HSCL-11, EPO-1, POMS). Study 1 asks whether deep learning outperforms classical machine learning: a two-stage deep pipeline, first predicting window-level valence/arousal and then session outcomes, transferred poorly (best single configuration, $r=.078$). Study 2 asks which representation and preprocessing choices work best: statistical pooling of frozen pretrained embeddings with regularized classical models achieved average $r=.406$ across nine scales under per-scale configuration selection, up to $r=.490$ for symptom distress, with complementary modalities and no significant benefit from manual affect annotations or therapist features. Study 3 asks whether deep learning remains inferior under these choices: classical models outperformed an optimized single-stage deep model at the single-configuration level ($r=.281$ vs.\ $.222$) and on all nine scales under matched per-scale selection ($.406$ vs.\ $.252$). The results suggest that under clinical data scarcity, pretrained representations with fixed statistical pooling and regularized classical models are more reliable than learned aggregation.
\end{abstract}


\section{Introduction}
\IEEEpubidadjcol

\IEEEPARstart{E}{motions} are a central component of psychotherapy, serving as both targets of therapeutic change and indicators of underlying clinical processes. A meta-analysis of 42 studies found a medium-to-large association between patients' emotional expressions during sessions and treatment outcome~\cite{peluso2018therapist}, and in-session affective experience, characterized by negative valence and high arousal, has been shown to predict subsequent symptom reduction~\cite{rubel2017patients}. Yet despite the established clinical relevance of emotional dynamics, their systematic assessment remains a major methodological bottleneck. Traditional approaches rely on post-session self-report questionnaires, which compress rich temporal processes into single retrospective judgments subject to recall bias~\cite{levine2002sources}, or on manual observer ratings, which are time-intensive and difficult to scale beyond small research samples~\cite{sharma2017continuous}.

Recent advances in artificial intelligence have begun to address this measurement gap. Video-based emotion recognition~\cite{terhurne2022validation}, sentiment analysis of therapy transcripts~\cite{eberhardt2025decoding}, and vocal affect detection~\cite{wagner2023emow2v} have each shown promise for capturing continuous emotional signals in naturalistic therapeutic settings. Prior automated work in psychotherapy has examined modalities largely in isolation and has often represented non-verbal channels by scalar affect outputs aggregated to coarse session-level summaries, leaving open whether the non-verbal channels lack outcome-relevant signal or whether the signal is lost in how their representations are extracted and aggregated. However, most automated emotion recognition systems are developed and validated on non-clinical datasets: acted performances, crowd-sourced recordings, or laboratory interactions. Such data differ substantially from psychotherapy in their affective characteristics. Therapy sessions involve subtle, often suppressed emotional expressions from patients with clinical disorders who may exhibit flattened affect, emotional incongruence across modalities, or patterns that deviate systematically from normative emotional displays~\cite{berenbaum1992emotional, gaebel2004facial}. This domain shift poses a fundamental challenge for transferring pretrained models to clinical applications.

The emergence of large-scale pretrained foundation models has reshaped the landscape of affective computing~\cite{schuller2026affective}. Self-supervised speech models such as Wav2Vec~2.0~\cite{baevski2020wav2vec} and its emotion-focused derivatives~\cite{wagner2023emow2v}, vision models trained on large-scale facial expression data~\cite{toisoul2021emonet}, and multilingual language models fine-tuned for sentiment analysis~\cite{barbieri2022sentiment} now provide rich, high-dimensional representations that encode affective information learned from vast corpora. These pretrained embeddings offer a compelling strategy for clinical applications: rather than training models from scratch on scarce clinical data, one can leverage representations learned from millions of non-clinical samples and adapt them to the therapeutic domain.

\begin{figure*}[t]
\centering
\resizebox{0.85\textwidth}{!}{%
\begin{tikzpicture}[
  font=\small, >=Stealth,
  node distance=22mm,
  io/.style={draw,rounded corners,align=center,minimum height=0.9cm,fill=gray!8},
  study/.style={draw,rounded corners,align=center,minimum height=2.6cm,text width=3.9cm,inner sep=3pt},
  good/.style={fill=green!12,draw=green!55!black},
  bad/.style={fill=red!8,draw=red!60},
  proc/.style={fill=blue!8,draw=blue!55!black},
  qline/.style={align=center,text width=3.9cm,font=\footnotesize\itshape},
  tlab/.style={midway,above,font=\footnotesize,align=center,fill=white,inner sep=2pt},
]
\node[io,text width=13cm] (rq)
  {\textbf{How can pretrained multimodal representations be leveraged for outcome prediction under data scarcity?}};
\node[study,good,below=16mm of rq] (s2)
  {\textbf{Study 2}\\[1mm]\footnotesize Statistical pooling of frozen embeddings $+$ classical ML; ablation over features, models, windows, slices\\[1.5mm]\normalsize $r = .406$ per-scale\\\normalsize $r = .490$ symptom distress};
\node[study,bad,left=22mm of s2] (s1)
  {\textbf{Study 1}\\[1mm]\footnotesize Two-stage deep learning with intermediate valence/arousal supervision\\[1.5mm]\normalsize $r = .078$};
\node[study,proc,right=22mm of s2] (s3)
  {\textbf{Study 3}\\[1mm]\footnotesize Single-stage deep learning incorporating Study-2 lessons\\[1.5mm]\normalsize $r = .222$ vs.\ $.281$\\\footnotesize single-config ML};
\node[qline,below=3mm of s1] (q1)
  {Does deep learning outperform classical ML under limited clinical data?};
\node[qline,below=3mm of s2] (q2)
  {Which representation and preprocessing choices are most effective?};
\node[qline,below=4mm of s3] (q3)
  {Does deep learning remain inferior under optimized choices?};
\draw[->] (rq.south) -- ++(0,-4mm) -| (s1.north);
\draw[->] (rq.south) -- (s2.north);
\draw[->] (rq.south) -- ++(0,-4mm) -| (s3.north);
\draw[->] (s1.east) -- node[tlab] {replace\\aggregation} (s2.west);
\draw[->] (s2.east) -- node[tlab] {re-test under\\Study-2 choices} (s3.west);
\end{tikzpicture}%
}
\caption{\textbf{Study roadmap.} Three studies answer the research question by comparing two-stage deep learning, statistical pooling of frozen pretrained embeddings with classical ML, and a simplified single-stage deep pipeline at $n = 106$ sessions. Classical ML on pooled pretrained embeddings outperforms both deep pipelines; numbers are average Pearson $r$ over nine clinical scales, with tiers (per-scale selection vs.\ single configuration) as labeled.}
\label{fig:concept}
\end{figure*}
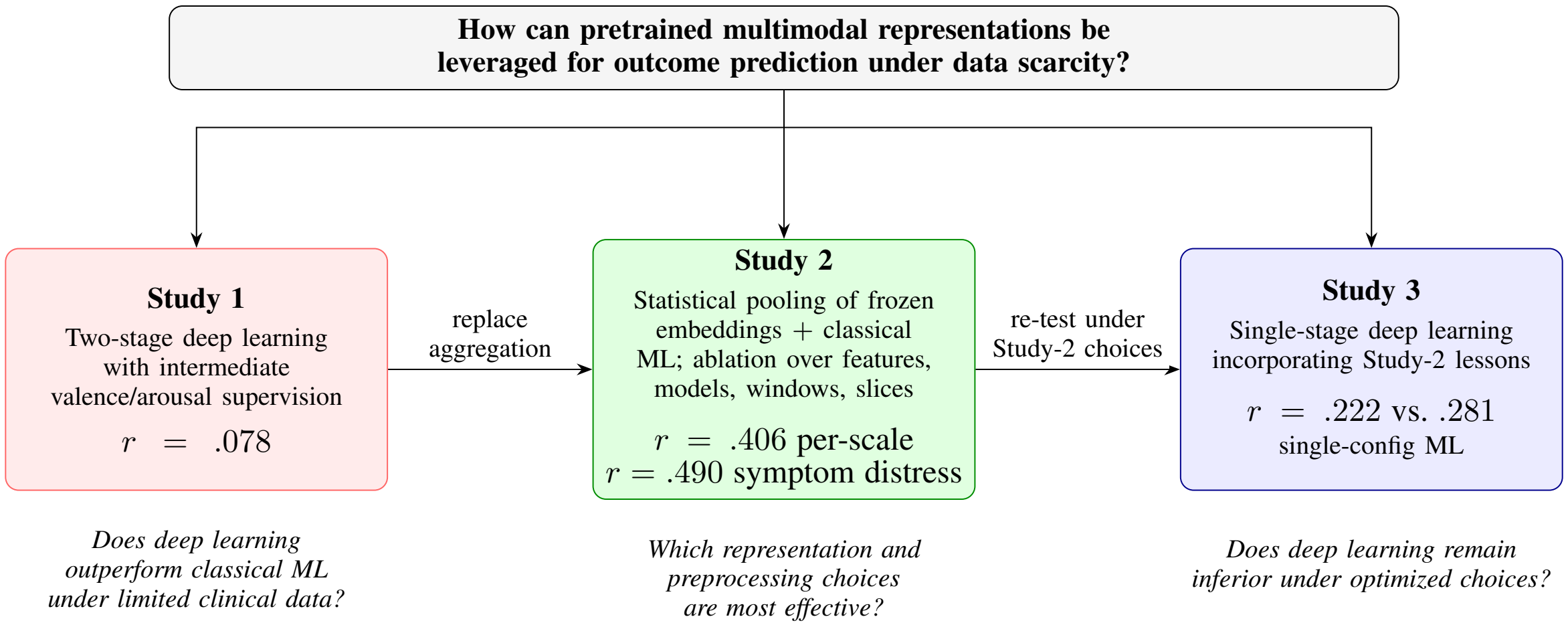

A critical open question, however, concerns how best to extract clinically relevant signal from these pretrained representations. The dominant paradigm in multimodal affective computing favors increasingly complex architectures: hierarchical temporal encoders, cross-modal attention mechanisms, end-to-end differentiable pipelines. They learn to fuse information across modalities and time~\cite{xiaoming2022survey}. While such approaches have achieved strong performance on benchmark datasets with tens of thousands of labeled samples, their suitability for clinical research settings is unclear. Psychotherapy datasets are inherently small due to privacy constraints, informed consent requirements, and the high cost of clinical data collection and annotation. Under these conditions, architectural complexity may become a liability rather than an asset, as models with many parameters are prone to overfitting when training data is limited.

Beyond statistical overfitting, end-to-end training on sensitive clinical data is also undesirable on privacy grounds: neural networks can memorize individual training samples~\cite{carlini2019secret, carlini2021extracting}, a concern that is acute for psychotherapy recordings, where disclosures are highly sensitive and the small number of patients makes individual sessions rare or unique examples. By decoupling frozen pretrained feature extraction from the downstream predictor rather than propagating questionnaire gradients into the encoders, our approach limits how much patient-specific information the trained model can absorb, complementing the local-processing data-protection measures (no cloud transmission) used in this study.

These privacy considerations align with a growing body of evidence from the machine learning literature demonstrating that tree-based and classical models consistently match or outperform deep learning on medium-sized tabular data~\cite{grinsztajn2022tree, shwartz2022tabular}. In clinical speech analysis, statistical pooling of pretrained speech embeddings has been shown to outperform more sophisticated temporal aggregation methods for Parkinson's disease detection~\cite{klempir2026statistical}. Similarly, in therapy outcome prediction from text, simpler Natural Language Processing (NLP) approaches have proven competitive with or superior to transformer-based architectures when applied to clinical datasets of realistic size~\cite{tanana2021you}.

In the present work, we investigate how pretrained multimodal representations can be most effectively leveraged for predicting clinical outcomes from psychotherapy sessions. We compare two fundamentally different strategies: (1)~temporal deep learning architectures that learn to aggregate frame-level embeddings through recurrent or attention-based mechanisms, and (2)~statistical feature extraction from the same embeddings followed by classical machine learning. Our approach integrates facial affect features from EmoNet~\cite{toisoul2021emonet}, vocal emotion representations from a Wav2Vec~2.0 model fine-tuned for dimensional emotion recognition (EmoW2V~\cite{wagner2023emow2v}), acoustic descriptors from eGeMAPS~\cite{eyben2015geneva}, and text-based sentiment from XLM-T~\cite{barbieri2022sentiment}. These multimodal features are extracted from 106 annotated psychotherapy sessions from 88 patients and used to predict standardized clinical questionnaires assessing symptom severity, emotional well-being, and in-session affect. Our design directly addresses this open question by testing whether richer representations, denser statistical pooling, and finer temporal granularity capture non-verbal signals that off-the-shelf affect summaries overlook; Figure~\ref{fig:concept} summarizes the three-study roadmap.

The contributions of this work are fourfold. First, to our knowledge, this is the first systematic comparison of temporal and attentional deep learning architectures with classical machine learning approaches for multimodal clinical outcome prediction from psychotherapy sessions, evaluated on identical patient-stratified folds. Second, we conduct a large-scale ablation study across prediction models, feature smoothing, session segmentation, modality subsets, and the inclusion of human annotations, comprising approximately 380,000 trained configurations in its first three steps. Third, we show that richer use of pretrained representations, combining embeddings with multi-statistic pooling at multiple temporal granularities, substantially improves multimodal prediction compared with coarse affect summaries. Fourth, a confirmatory contrast indicates that continuous human affect annotations provide no significant additional predictive value, and therapist-side features did not improve prediction on any scale.

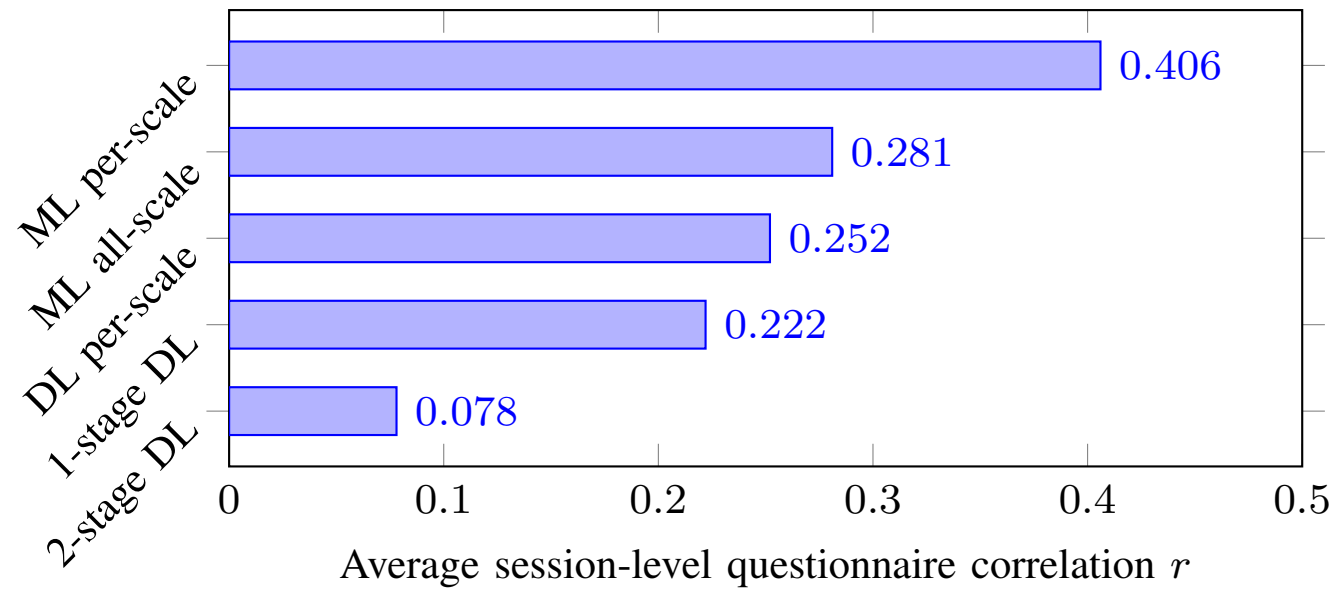
\begin{figure}[t]
\centering
\begin{tikzpicture}
\begin{axis}[
    xbar,
    width=0.98\linewidth, height=4.6cm,
    xmin=0, xmax=0.5,
    xlabel={Average session-level questionnaire correlation $r$},
    symbolic y coords={2-stage DL,1-stage DL,DL per-scale,ML all-scale,ML per-scale},
    ytick=data,
    yticklabel style={rotate=45,anchor=east,font=\footnotesize},
    nodes near coords,
    nodes near coords align={horizontal},
    every node near coord/.append style={
        font=\footnotesize,
        /pgf/number format/fixed,
        /pgf/number format/precision=3,
    },
    enlarge y limits=0.16,
    bar width=9pt,
    tick label style={font=\footnotesize},
    label style={font=\footnotesize},
]
\addplot coordinates {
    (0.078,2-stage DL)
    (0.222,1-stage DL)
    (0.252,DL per-scale)
    (0.281,ML all-scale)
    (0.406,ML per-scale)};
\end{axis}
\end{tikzpicture}
\caption{\textbf{Headline result.} Average session-level questionnaire correlation across nine clinical scales. The two-stage deep baseline is near zero. The simplified single-stage deep model improves substantially but trails single-configuration ML ($.222$ vs.\ $.281$); per-scale selection widens the gap ($.406$ vs.\ DL's per-scale oracle $.252$).}
\label{fig:headline}
\end{figure}

Our key finding is that frozen pretrained embeddings, summarized by simple session statistics and fed to classical machine learning, suffice for clinically meaningful outcome prediction from roughly one hundred sessions, outperforming every temporal deep learning architecture we tested at a fraction of the training cost (Figure~\ref{fig:headline}). Through a systematic ablation study, we identify optimal configurations across the key design dimensions of this prediction task. The resulting models achieve meaningful associations with session-level clinical assessments, demonstrating that pretrained multimodal representations contain clinically relevant information that can be extracted through computationally efficient means.

These findings contribute to both the affective computing and psychotherapy research literatures. From a methodological perspective, they challenge the assumption that complex temporal architectures are necessary for clinical prediction from pretrained embeddings, and provide evidence-based guidance for model selection under data scarcity. From a clinical perspective, they demonstrate a practical, privacy-conscious approach for linking continuous multimodal emotional dynamics to validated clinical assessments in naturalistic therapy settings, with implications for process-outcome research and measurement-based care.

Feature extraction relies on publicly available DISCOVER modules~\cite{hcmlab_discover_modules}; the pooling, training, cross-validation, and per-scale model-selection code is released at \url{https://github.com/saveli/psychotherapy-outcome-fusion}. Trained models are not released owing to consent restrictions on the underlying clinical data.

Guided by the research question, \emph{How can pretrained multimodal representations be most effectively leveraged for psychotherapy outcome prediction under data scarcity?}, the paper proceeds through three studies. Study 1 asks whether deep learning outperforms classical ML under limited clinical data. Study 2 asks which representation and preprocessing choices are most effective when frozen pretrained facial, vocal, and textual embeddings are statistically pooled and paired with regularized classical models. Study 3 returns to deep learning not as a repetition of Study 1, but as a fair re-test under the representation and preprocessing choices optimized in Study 2, asking whether deep learning remains inferior once those choices are held to their strongest observed form.

\section{Related Work}

\subsection{Emotion Measurement in Psychotherapy Process-Outcome Research}

The relationship between emotional processes during therapy and clinical outcomes has been a longstanding focus of psychotherapy research. Emotions serve multiple functions in treatment: they signal patients' underlying needs and motives, they are transdiagnostically relevant across therapeutic modalities~\cite{whelton2004emotional}, and their expression is associated with outcome across a range of approaches~\cite{peluso2018therapist}. In-session affective experience, particularly the combination of negative valence with high arousal that characterizes emotionally engaging therapeutic work, has been identified as a predictor of session-to-session symptom change~\cite{rubel2017patients, feldman1995valence}.

Traditional measurement of these processes relies on self-report instruments administered before or after sessions (e.g. the POMS~\cite{mcnair1992edits} for session-level affect, the HSCL-11~\cite{lutz2006reliabilitat} for symptom distress) or on observer-based rating systems that require trained coders to review recordings. While clinically informative, these approaches face well-documented limitations: retrospective self-reports are subject to memory biases~\cite{levine2002sources}, observer ratings are time-intensive and expensive to scale, and both provide only coarse temporal resolution compared to the moment-to-moment emotional dynamics that unfold within sessions.

Automated approaches have begun to address these limitations, but they differ fundamentally in how they represent multimodal information; Table~S1 in the supplementary material\footnote{Supplementary material is available in the online version.} summarizes them by representation type. One line uses classical or handcrafted affect features. \citet{terhurne2022validation} used NOVA~\cite{Baur2020} to train video-based emotion recognition models on OpenFace facial features~\cite{baltruvsaitis2016openface}, including action-unit-based facial information, for continuous prediction of valence and arousal in therapy sessions. On the 106 annotated sessions reused in the present work, their models achieved moderate agreement with human raters
and showed significant associations with therapist-rated emotions and the anxiety subscale of the HSCL-11, establishing a proof of concept for video-based affect assessment in naturalistic psychotherapy.

A second line represents sessions through scalar outputs of pretrained affect or sentiment models, typically summarized to coarser session-level statistics. \citet{eberhardt2025decoding} extended automated assessment to the text modality, using a transformer-based sentiment analysis model to assess emotional tone in therapy transcripts. The representation here is the model-derived textual sentiment signal rather than a multimodal embedding. The study found significant correlations with both self-reported and therapist-reported emotions and predictive associations with treatment outcome. More broadly, the systematic review by \citet{malgaroli2023natural} identified growing adoption of machine learning for analyzing therapy conversations, with text as the dominant modality and NLP representations varying across the reviewed studies, and noted that multimodal approaches remain rare.

A third possibility is to use pretrained embeddings themselves as the representation. To our knowledge, prior psychotherapy outcome work has not tested embedding-level multimodal fusion across facial, vocal, and textual streams. The present work fills this gap by combining facial features from EmoNet, vocal emotion representations from EmoW2V, handcrafted eGeMAPS acoustic descriptors, and text-based sentiment from XLM-T, including pretrained embeddings where available. We compare these richer representations with scalar affect summaries, apply denser statistical pooling across multiple smoothing windows and session segments, and systematically compare learned against fixed aggregation strategies for predicting nine session-level questionnaire scales.

The special issue on multimodal measurement in \textit{Psychotherapy Research}~\cite{doorn2025development, aafjes2025intuition} has highlighted both the promise and the methodological challenges of integrating multiple data streams for studying therapeutic processes, particularly the need for transparent, explainable, and well-calibrated methods.

\subsection{Pretrained Foundation Models for Affective Computing}

The field of affective computing has undergone a paradigm shift with the emergence of large-scale pretrained models~\cite{schuller2026affective}. Where earlier approaches relied on handcrafted features (e.g. spectral functionals for speech, muscle group activations for facial expressions, bag-of-words representation for text) fed into conventional classifiers, current methods increasingly leverage high dimensional embeddings learned through self-supervised pretraining on massive unlabeled corpora.

In the speech domain, Wav2Vec 2.0~\cite{baevski2020wav2vec} and HuBERT~\cite{hsu2021hubert} have emerged as foundation models for a range of downstream tasks. Wagner et al.~\cite{wagner2023emow2v} conducted a comprehensive evaluation of these architectures for dimensional speech emotion recognition, fine-tuning several model variants on the MSP-Podcast corpus~\cite{Busso_2025} for predicting arousal, dominance, and valence. Their work demonstrated that self-supervised pretraining is critical for capturing affective information, particularly for valence, where a randomly initialized Wav2Vec~2.0 model performed substantially worse than even a standard Convolutional Neural Network (CNN) baseline. The resulting model (EmoW2V~\cite{wagner2023emow2v}), which we employ in the present study, achieved state-of-the-art concordance correlation coefficients for valence prediction and showed meaningful cross-corpus generalization.

For the visual modality, EmoNet~\cite{toisoul2021emonet} provides continuous estimates of valence, arousal, and categorical emotions from facial images, trained on the AffectNet~\cite{mollahosseini2017affectnet} dataset. In the text domain, transformer-based multilingual models fine-tuned for sentiment analysis, such as XLM-T~\cite{barbieri2022sentiment} and the broader family of XLM-RoBERTa models~\cite{conneau2020unsupervised}, offer robust cross-lingual sentiment representations.

A central insight from this literature is that pretrained models encode rich affective information even when not explicitly trained for emotion recognition. \citet{pepino2021emotion} demonstrated that simple neural networks applied to frozen Wav2Vec~2.0 embeddings could achieve competitive speech emotion recognition, suggesting that the pretrained representations already capture much of the relevant structure. This raises a question that motivates our work: if pretrained embeddings already encode affective content, how much additional architectural complexity is needed for downstream clinical prediction tasks?

\subsection{Multimodal Fusion for Emotion Recognition}

Multimodal emotion recognition integrates complementary information from visual, acoustic, and linguistic channels, motivated by the observation that human emotional communication is inherently multimodal~\cite{baltruvsaitis2018multimodal}. Fusion strategies are commonly categorized as early (feature-level), late (decision-level), or hybrid~\cite{xiaoming2022survey}. Early fusion concatenates features before modeling, preserving cross-modal interactions but increasing dimensionality. Late fusion combines independent unimodal predictions, offering flexibility and modularity but forgoing inter-modal dependencies. Hybrid approaches, including attention-based mechanisms and cross-modal transformers, aim to combine the strengths of both~\cite{tsai2019multimodal}.

Recent trends in the field have pushed toward increasingly sophisticated fusion architectures. Cross-modal attention mechanisms learn to align and weight information across modalities dynamically. Hierarchical temporal models capture affective dynamics at multiple time scales. End-to-end differentiable pipelines jointly optimize feature extraction and fusion. These approaches have achieved strong results on benchmark datasets such as IEMOCAP~\cite{busso2008iemocap}, MELD~\cite{poria2019meld}, and CMU-MOSI/MOSEI~\cite{zadeh2018multimodal}, which typically provide thousands of labeled samples across controlled recording conditions.

However, the applicability of these complex architectures to clinical settings has received limited attention. Psychotherapy data differs from standard benchmarks in several critical respects: sample sizes are constrained by ethical and practical considerations, emotional expressions are subtle and naturalistic rather than exaggerated or acted, sessions are long (typically 50 minutes) rather than short utterances, and the prediction targets are clinical questionnaires rather than categorical emotion labels. These characteristics may favor simpler, more robust approaches over complex architectures that require large training sets to avoid overfitting. The present work tests this hypothesis directly by comparing temporal and attentional deep learning architectures against statistical pooling with classical machine learning on the same clinical data.

\subsection{Pretrained Embeddings with Classical Machine Learning vs.\ Deep Learning Under Data Constraints}

A substantial body of evidence has accumulated showing that tree-based ensemble methods remain competitive with or superior to deep learning for tabular prediction tasks, particularly on medium-sized datasets. \citet{grinsztajn2022tree} benchmarked standard and novel deep learning methods against XGBoost~\cite{chen2016xgboost} and Random Forests~\cite{breiman2001rf} across 45 datasets, finding that tree-based models were state-of-the-art on datasets with approximately 10,000 samples, even without accounting for their faster training times. \citet{shwartz2022tabular} reached similar conclusions, showing that XGBoost outperformed recently proposed deep tabular models across diverse datasets while requiring substantially less hyperparameter tuning.

These findings have direct implications for clinical applications where data scarcity is the norm rather than the exception. In clinical speech analysis, a recent study on Parkinson's disease detection from speech found that statistical pooling of Wav2Vec~2.0 embeddings combined with classical classifiers outperformed more sophisticated temporal aggregation approaches~\cite{klempir2026statistical}, a finding that closely parallels our own observations in the psychotherapy domain. In therapy outcome prediction from text, \citet{tanana2021you} found that a simple unigram sentiment model outperformed fine-tuned BERT~\cite{DBLP:journals/corr/abs-1810-04805} for automatically rating emotion in psychotherapy sessions, suggesting that the advantages of complex models diminish when training data is limited and the domain is specialized.

The pattern extends beyond tabular data. In the broader machine learning literature, \citet{amin2023will} evaluated ChatGPT (built on the GPT-3 model family~\cite{brown2020language}) on affective computing tasks including personality prediction and sentiment analysis, finding that a task-specific fine-tuned RoBERTa~\cite{liu2019roberta} model generally outperformed the much larger general-purpose model. This underscores a recurring theme: for specialized downstream tasks with limited data, targeted use of pretrained representations with simple prediction heads often outperforms more complex end-to-end approaches.

Our work contributes to this literature by providing systematic evidence from the clinical psychotherapy domain, comparing temporal and attentional deep learning architectures against classical machine learning across multiple modalities, fusion strategies, normalization regimes, and clinical outcome measures on identical patient-stratified folds. To our knowledge, this is the first study to conduct a systematic deep-learning-versus-classical-ML comparison for multimodal clinical outcome prediction from psychotherapy session data.

\section{Dataset and Feature Extraction}
\subsection{Sample characteristics}
A total of 106 sessions drawn from 88 patients and 53 therapists were analyzed for model development. Of the patients, 62.5\% were female, and the mean age was 35.36 years (\textit{SD} = 14.70). On average, treatments comprised 31.88 sessions (\textit{SD} = 17.07), with a dropout rate of 12.5\%. With respect to primary diagnoses, affective disorders were most frequent (47.73\%), followed by anxiety disorders (13.64\%), adjustment disorders (13.64\%), posttraumatic stress disorder (7.95\%), somatization disorder (5.68\%) and other diagnoses (11.36\%).

\begin{figure}[!t]
    \includegraphics[width=\columnwidth]{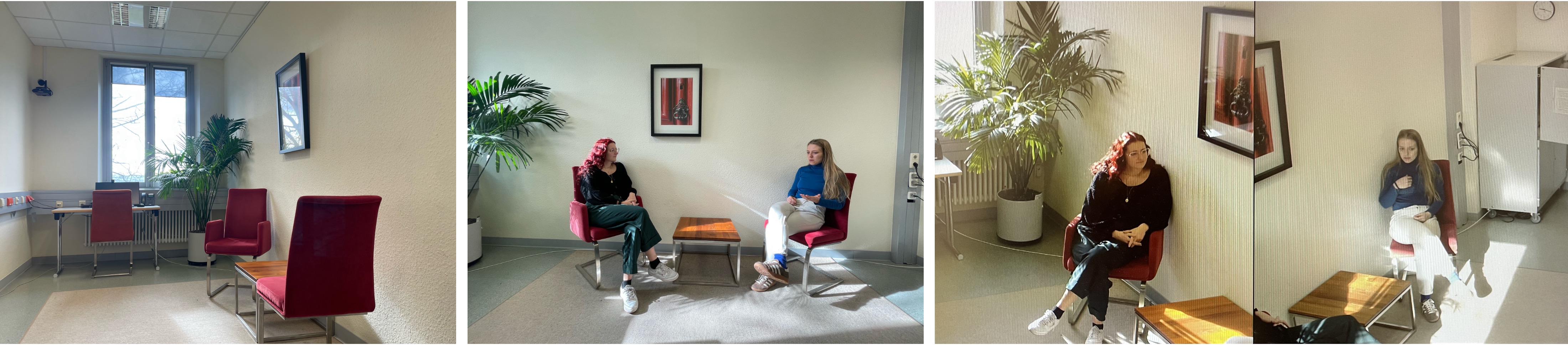}
    \caption{\textbf{Therapy recording setup and multimodal input.} \textit{Left:} Empty therapy room with two corner-mounted cameras, one ceiling-mounted microphone, and the routine outcome monitoring (ROM) computer at the rear of the room. \textit{Middle:} Example naturalistic therapy session with therapist and patient seated opposite each other. \textit{Right:} Synchronized therapist- and patient-facing video streams, together with a shared audio stream, used as input to the analysis pipeline.}
    \label{fig:therapieraum}
\end{figure}

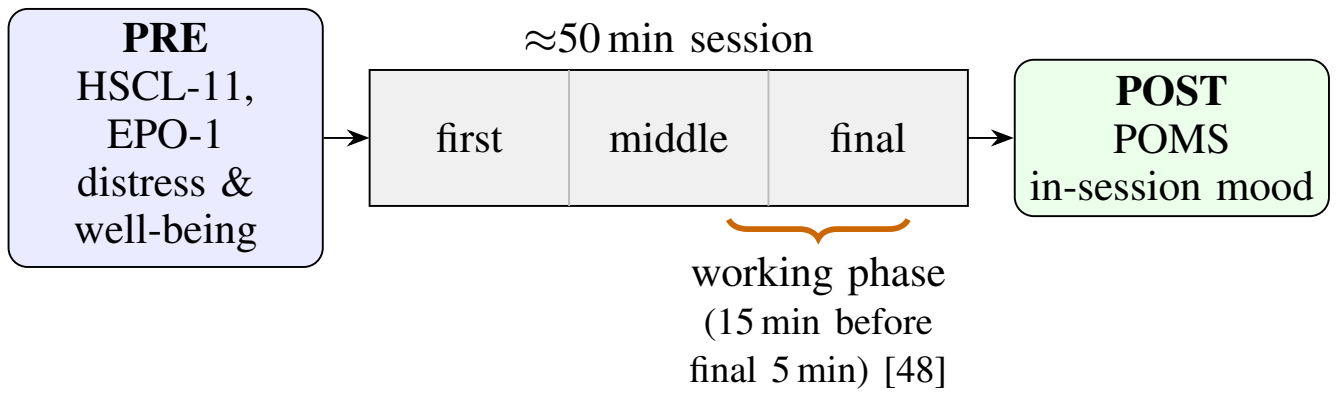
\begin{figure}[t]
\centering
\resizebox{\columnwidth}{!}{
\begin{tikzpicture}[
  font=\small, >=Stealth,
  box/.style={draw,rounded corners,align=center,minimum height=1.15cm,text width=2.1cm,inner sep=3pt},
]
\def\W{4.4}
\def\X{1.5}
\node[box,fill=blue!8] (pre) at (0,0) {\textbf{PRE}\\HSCL-11, EPO-1\\distress \& well-being};
\filldraw[fill=gray!12,draw=black] (\X,-0.5) rectangle ++(\W,1.0);
\draw[gray!55] (\X+\W/3,-0.5)   -- ++(0,1.0);
\draw[gray!55] (\X+2*\W/3,-0.5) -- ++(0,1.0);
\node at (\X+\W/6,0)   {first};
\node at (\X+\W/2,0)   {middle};
\node at (\X+5*\W/6,0) {final};
\node[above] at (\X+\W/2,0.5) {$\approx$50\,min session};
\draw[decorate,decoration={brace,amplitude=4pt,mirror},very thick,orange!80!black]
  (\X+0.60*\W,-0.6) -- (\X+0.90*\W,-0.6);
\node[below,text width=3.0cm,align=center] at (\X+0.75*\W,-0.75)
  {working phase\\\footnotesize (15\,min before final 5\,min)~\cite{auszra2013client}};
\node[box,fill=green!8] (post) at (\X+\W+1.5,0)
  {\textbf{POST}\\POMS\\in-session mood};
\draw[->] (pre.east) -- (\X,0);
\draw[->] (\X+\W,0) -- (post.west);
\end{tikzpicture}%
}
\caption{\textbf{Session structure and measurement timing.} Pre-session questionnaires capture symptom distress and well-being; post-session questionnaires capture in-session mood. The session is sliced into thirds and a working-phase segment (the 15 minutes preceding the final 5 minutes) for the segmentation ablation (Study~2).}
\label{fig:session}
\end{figure}

The study was conducted in accordance with the Declaration of Helsinki and was approved by the Ethics Committee of the German Psychological Society (DGPs). Prior to the start of therapy, all patients were informed about the study procedures and the use of their data for research purposes. Participation in the study was voluntary. All participants were asked to provide written informed consent. Patients could decline participation or withdraw consent at any time without any consequences for their treatment. Protecting patients’ confidentiality and privacy was a core objective of the study. Video recordings and derived session transcripts were handled confidentially and stored on secure institutional servers. Automated transcription and any subsequent model-based analyses were performed locally using models deployed on institutional infrastructure. Importantly, recordings and transcripts were not transmitted to third-party providers, cloud-based services, or external APIs. The participants of this study did not give written consent for their data to be shared publicly. Thus the sensitive nature of the research does not allow for the data to be made available.
\subsection{Treatment and Recording Setup}

The data used for the study were collected from therapy sessions in personalized integrative cognitive-behavioral therapy (CBT) at a university-based outpatient training and research clinic in Southwest Germany. Treatment was delivered in individual weekly sessions lasting around 50 minutes. Therapists were trained in manualized CBT interventions and transtheoretical concepts and received regular supervision. Supervision was provided by licensed CBT clinicians with at least five years' experience in clinical practice and advanced training in clinical supervision.
As part of routine clinical procedures, all sessions were recorded on video using Telycam TLC-700-S-R cameras and Beyerdynamic BM 32 W microphones. The recorded videos were standardized to a resolution of $640 \times 480$ pixels and 25 frames per second. Audio was recorded in mono at a sample rate of 48\,kHz. The recording setup is shown in Fig.~\ref{fig:therapieraum}. Questionnaire batteries were administered as part of routine outcome monitoring with an implemented feedback system used by the therapists. For this study, questionnaires completed by patients and therapists at session level were used. Patients filled out the questionnaires before and after the session, while therapists filled out the questionnaire only at the end.

\subsection{Questionnaires} \label{sec:Questionaires}
Figure~\ref{fig:session} summarizes when each instrument is administered.
\textbf{HSCL-11.} The Hopkins Symptom Checklist-11 (HSCL-11; \cite{lutz2006reliabilitat}), an 11-item self-report questionnaire, was used to assess patient symptomatic distress at the beginning of each session. Patients rated the extent to which they experienced each symptom during the previous seven days on a 4-point Likert scale ranging from 1 (\textit{not at all}) to 4 (\textit{extremely}). The items primarily assess symptoms of depression and anxiety. The mean score reflects the patient’s overall level of symptom distress for the preceding week, while separate subscale scores can be calculated for anxiety- and depression-related symptoms. The global score demonstrates high internal consistency (\(\alpha = .92\)) and shows a strong correlation (\textit{r} = .91) with the Brief Symptom Inventory (BSI \cite{derogatis1992scl}) \cite{lutz2006reliabilitat}.

\textbf{EPO-1.} General psychological well-being was assessed with the Emotional and Psychological Outcome item (EPO-1), a single-item measure used for routine outcome monitoring. Patients rate their current emotional and psychological functioning (\textit{“At this moment, how well do you feel you are getting along emotionally and psychologically?”}) at the beginning of each session. Two response formats are reported: a 5-point Likert scale ranging from 0 (\textit{very poorly}) to 4 (\textit{very well}), adopted by the European Psychotherapy Consortium (EPoC) as a common outcome measure \cite{gonccalves2024developing}, and a 0--100 visual analog scale (VAS) enabling more fine-grained assessment \cite{lutz2019towards}. Both formats correlate with established symptom/outcome measures and are sensitive to change during psychotherapy \cite{lutz2021measuring}. In the present study, both formats were analyzed and compared.

\textbf{POMS.} The emotional experiences of patients during sessions were assessed using seven items from the Profile of Mood States (POMS \cite{mcnair1992edits}). At the end of each session, the therapist and the patient rated the extent to which the patient felt \textit{sad}, \textit{ashamed}, \textit{anxious}, \textit{angry}, \textit{content}, \textit{energetic} and \textit{relaxed} during the session, using a scale ranging from 0 (\textit{not at all}) to 100 (\textit{extremely}). Ratings for sadness, shame, anxiety and anger were averaged to produce a negative emotion score, calculated separately for the patient and therapist. Similarly, ratings for contentment, energy and relaxation were averaged to produce separate positive emotion scores.

\subsection{Human annotations} \label{sec:humananno}
Human annotations were conducted to enable continuous measurement of the affective dimensions valence and arousal during each therapy session (see \citet{terhurne2022validation}). Both dimensions were operationalized on a scale ranging from $-1$ (\textit{very negative} / \textit{very low arousal}) to $+1$ (\textit{very positive} / \textit{very high arousal}), with $0$ defined as a neutral baseline. Ratings were restricted to the therapeutically relevant core content of each session, explicitly excluding introductory and closing small talk. For each session, valence was annotated first, followed by arousal. While watching the videos with audio, raters continuously adjusted their annotations in real time by moving the mouse wheel up or down to indicate changes in valence or arousal, with the option to pause, navigate to specific timestamps, and correct previous entries. In cases of discrepancies between verbal content and non-verbal emotional expression, observable affective cues (e.g. facial expressions, vocal tone) were prioritized. All four raters (two female, two male) were psychology graduates who completed 30 hours of structured training, including technical instruction, foundations of emotional theory with a focus on the circumplex model~\cite{russell1980circumplex}, and supervised practice using example videos. Raters began contributing to the dataset only after the rater pool reached an inter-rater reliability of at least Cronbach’s \(\alpha = .70\). The final annotations reached average inter-rater reliabilities of \(\alpha = .74\) for valence (range $.60$--$.90$) and \(\alpha = .73\) for arousal (range $.61$--$.89$)~\cite{terhurne2022validation}. All raters received monthly supervision from a licensed psychotherapist.

\subsection{Feature extraction} \label{sec:featureextrac}
All three studies operate on the same frame-level multimodal features extracted from the raw session recordings. All extraction is performed using the DISCOVER~\cite{hallmen2025discover} framework and its ready-to-run modules~\cite{hcmlab_discover_modules}, of which we describe the applied ones in more detail:

\textbf{Text.} To derive text-based automatic measures from therapy sessions, session audio was converted into time-aligned transcripts and sentiment was extracted as a predictor for subsequent modeling. Speech-to-text transcription was performed with WhisperX~\cite{bain2023whisperx}, an adaptation of Whisper~\cite{radford2023robust}, which provides improved timestamp accuracy and supports long recordings. Because many sessions were recorded as a single mixed audio channel, speaker diarization was applied to assign transcript segments to individual speakers. Diarization was carried out using \textit{pyannote}~\cite{bredin2020pyannote} and \textit{SpeechBrain}~\cite{ravanelli2024open} by embedding voiced segments and clustering them via unsupervised agglomeration. Clusters were then mapped to patient vs. therapist using an oracle-based procedure with reference speech segments. Sentiment was estimated from the resulting transcripts using \textit{XLM-T}~\cite{barbieri2022sentiment}. Both the model’s continuous sentiment score, ranging from $-1$ (\textit{negative}) through $0$ (\textit{neutral}) to $+1$ (\textit{positive}), and the corresponding text embeddings were extracted as predictors.

\textbf{Audio.} To derive audio-based automatic measures, we extracted frame-wise vocal features from session recordings using two complementary approaches. First, we applied a pretrained vocal emotion detection model (EmoW2V~\cite{wagner2023emow2v}) to obtain affect-focused representations. From \textit{EmoW2V}, we extracted both its dimensional affect estimates (e.g. \textit{valence}, \textit{arousal}, and \textit{dominance}) and the corresponding embedding representations as predictors for subsequent modeling. Second, we computed a broader set of handcrafted acoustic descriptors
using the \textit{openSMILE}~\cite{eyben2010opensmile} toolkit. Specifically, we extracted the \textit{eGeMAPSv02}~\cite{eyben2015geneva} feature set, which provides a standardized, general-purpose baseline for speech-related prediction and affective computing. All audio-derived features were extracted frame-wise, i.e.\ 40\,ms windows for 25\,fps video, to enable consistent smoothing in later stages.

\textbf{Video.} To derive video-based automatic measures, we extracted frame-wise facial affect features from session recordings using a facial emotion recognition model (EmoNet \cite{toisoul2021emonet}). From each video frame, we obtained the model’s continuous affect estimates (\textit{valence} \& \textit{arousal}), its categorical emotion prediction, and the corresponding embedding representations as predictors for subsequent modeling.

\section{Study 1: Two-Stage Deep Learning with Intermediate Valence/Arousal Supervision}

Study 1 asks whether deep learning outperforms classical machine learning under limited clinical data. Using the shared multimodal features (Section~\ref{sec:featureextrac}), it establishes a two-stage baseline that first predicts continuous valence/arousal under intermediate supervision and then maps these estimates to session-level questionnaires, with classical ML evaluated on identical patient-stratified folds.

\subsection{Method}

\begin{figure}[!t]
    \centering
    \includegraphics[width=\linewidth]{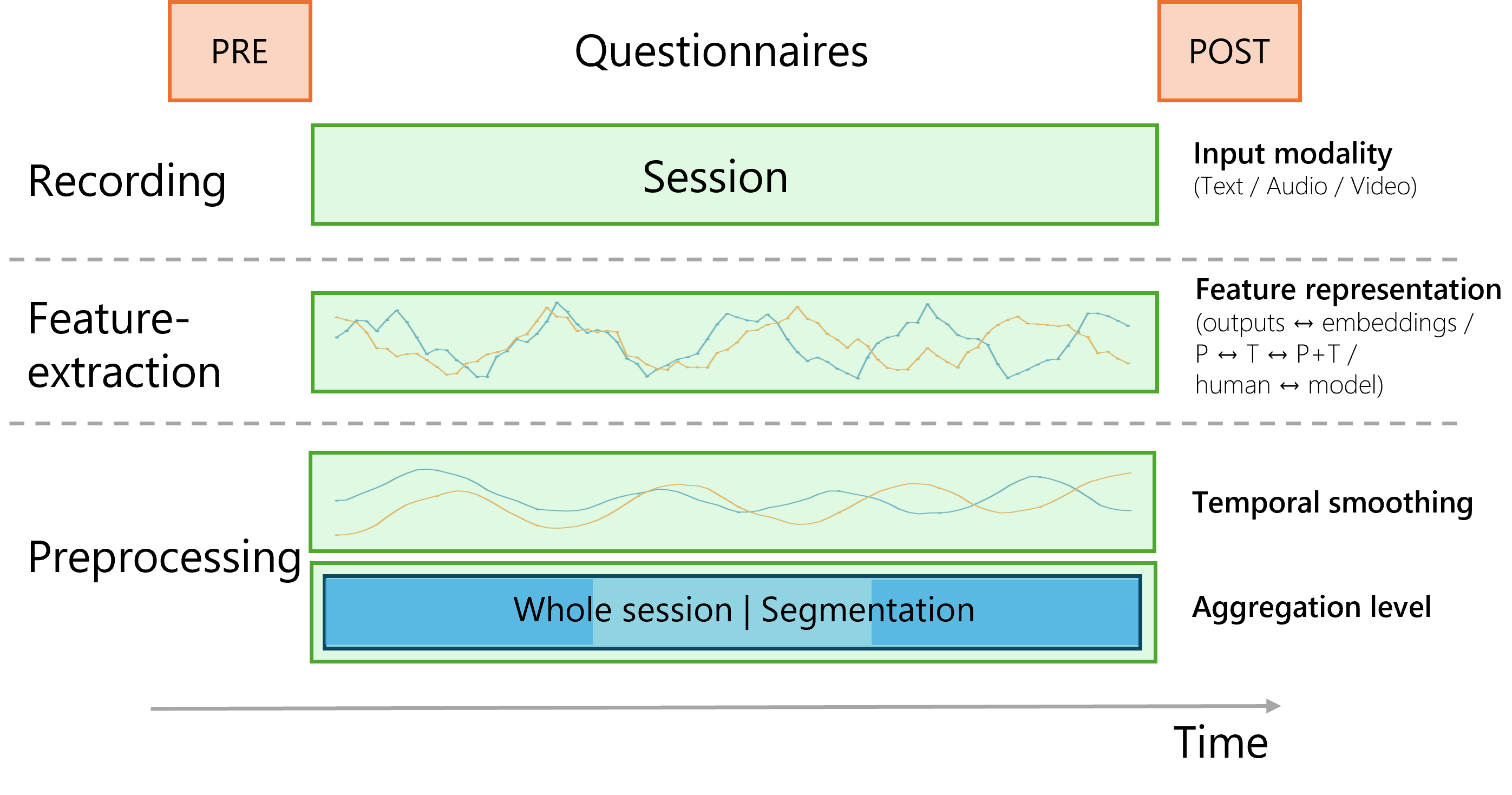}
    \caption{\textbf{Overview of the multimodal affective prediction pipeline.} The left panel lists the core development steps, while the right panel specifies the corresponding modeling and preprocessing degrees of freedom varied across ablations. Pre-session questionnaires (PRE) quantify patients’ current symptoms, whereas post-session questionnaires (POST) target session processes (i.e. process-based measures of in-session experience and interaction). Continuous valence–arousal estimates are derived from session data, optionally smoothed, and summarized at different temporal levels (whole session vs. segments) to optimize prediction of questionnaire values.}
    \label{fig:baseline}
\end{figure}

\subsubsection{Performance targets}

We treat the two-stage, annotation-supervised pathway of \citet{terhurne2022validation}, which first predicts continuous affect annotations and then relates these estimates to clinical questionnaires, as our baseline. Rather than treating higher frame-level valence/arousal accuracy as a goal in itself, we ask (1) whether multimodal pretrained features with learned temporal aggregation yield session-level questionnaire associations of practical magnitude, and (2) whether the labor-intensive continuous valence/arousal annotation, and the intermediate-supervision pipeline it enables, is justified by its contribution to that prediction. Study~1 establishes the baseline; Studies~2--3 answer both.

\subsubsection{Prediction pipeline}

Clinical outcome prediction from therapy sessions requires bridging two temporal scales: frame-level affective signals extracted at 25\,fps and session-level questionnaire scores collected once per session. We decompose this into two stages and systematically vary the architectural choices within each (cf.\ Figure~\ref{fig:baseline}).

\subsubsection{Stage 1: Frame-to-window affect prediction}
\label{sec:stage1}

In the first stage, the goal is to predict continuous valence and arousal (V/A) from multimodal input features at the temporal resolution of the continuous V/A annotations of \citet{terhurne2022validation}, which we reuse as Stage-1 supervision targets. We construct overlapping windows of 50 frames (2\,s) with a stride of 25 frames (50\% overlap), yielding approximately 307{,}000 training windows across the 106 annotated sessions. The window allows the models to adapt to the inherent smoothing and delay of human continuous real-time annotation.

\textbf{Input features.} For each window, we construct a 7-dimensional frame-level affect feature vector from the modalities introduced in Section~\ref{sec:featureextrac}: three EmoNet-derived facial features (emotion, valence, arousal), three EmoW2V-derived vocal features (valence, arousal, dominance), and one XLM-T-derived textual feature (sentiment). The handcrafted eGeMAPS descriptors are not part of this affect vector; they enter the feature pool only in the Study~2 ablation. Per-patient z-score normalization of these features was evaluated to reduce between-subject variability but degraded performance, most severely for the deep learning pipelines, so all reported Study~1 results use unnormalized features.

A second configuration additionally includes the pretrained embeddings (EmoNet 256-d, EmoW2V 1024-d, XLM-T 768-d), trading interpretability and dimensionality for a richer representation.

\textbf{Encoder architectures.} We compare three encoder families for predicting V/A from the windowed features: classical ML models and two deep learning sequence encoders:

\begin{itemize}
    \item \textbf{Classical ML:} Ridge~\cite{hoerl1970ridge}, XGBoost~\cite{chen2016xgboost}, Random Forest~\cite{breiman2001rf}, and a shallow MLP are trained to predict V/A independently for each frame within a window, without modeling temporal dependencies.
    \item \textbf{LSTM:} A two-layer LSTM~\cite{hochreiter1997lstm} with hidden size 8 processes the window as a sequence, producing a V/A prediction at each time step. This architecture captures local temporal dynamics within the 2-second context.
    \item \textbf{Transformer:} A transformer encoder with learned positional embeddings processes the window in parallel, attending to all frames simultaneously. This tests whether self-attention captures frame-level affect patterns more effectively than recurrent processing.
\end{itemize}

\subsubsection{Stage 2: Session-level questionnaire prediction}
\label{sec:stage2}

In the second stage, frame- or window-level representations are aggregated across the full session and used to predict clinical questionnaire scores. The key architectural choice at this stage is how to perform the temporal aggregation from frame-level to session-level.

\textbf{Aggregation strategies.} We compare three approaches:

\begin{itemize}
    \item \textbf{Statistical pooling:} For each feature dimension, five summary statistics are computed across the session: mean, standard deviation, minimum, maximum, and median. With $d$ input features, this produces a $5d$-dimensional session-level vector. This approach discards temporal ordering entirely but produces a fixed-size representation regardless of session length.
    \item \textbf{Attention pooling:} A learned attention mechanism computes a weighted average over frame-level representations, allowing the model to attend selectively to time steps that are most informative for the clinical prediction target.
    \item \textbf{LSTM aggregation:} A two-layer LSTM with hidden size 16 processes the sequence of frame-level representations across the entire session, using its final hidden state as the session-level representation. This approach learns temporal aggregation implicitly.
\end{itemize}

\textbf{Prediction models.} Session-level representations are mapped to questionnaire scores using one of several prediction models. For configurations using statistical pooling, the resulting tabular features are compatible with both classical ML and neural approaches. We evaluate: Ridge Regression~\cite{hoerl1970ridge},
Support Vector Regression (SVR~\cite{drucker1997svr}), Gaussian Process Regression (GPR~\cite{rasmussen2006gpr}), Random Forest~\cite{breiman2001rf}, XGBoost~\cite{chen2016xgboost}, Gradient Boosted Decision Trees (GBDT~\cite{friedman2001gbdt}), and a two-layer MLP with separate prediction heads per questionnaire, dropout of $0.2$, and hidden dimensions of 128 and 64. For configurations using attention or LSTM aggregation, the aggregation module and prediction head are trained jointly via back-propagation.

\textbf{Therapist features.} In all configurations described above, only patient features are used as input. As an additional condition, we include the same multimodal features extracted from the therapist stream. Since no V/A annotations exist for the therapist, therapist features bypass the first stage entirely and enter the pipeline only at the session-level prediction stage. This allows the model to capture therapist behavior and implicit patient--therapist dynamics that may influence session-level outcomes.

\subsubsection{Training strategies}
\label{sec:training}

The two-stage decomposition permits several training configurations that we compare systematically:

\begin{itemize}
    \item \textbf{Two-stage:} Stage~1 and Stage~2 are trained independently. The frame-level encoder is optimized for V/A prediction. Its outputs or intermediate representations are then used as fixed inputs for the session-level model. This decoupling is the default configuration for all classical ML approaches.
    \item \textbf{End-to-end:} Gradients from the session-level questionnaire loss are propagated through the aggregation module and the frame-level encoder, jointly optimizing both stages. This allows the encoder to learn representations tuned for downstream clinical prediction rather than solely for V/A reconstruction but still leveraging the intermediate human annotations.
    \item \textbf{Curriculum learning:} A variant of end-to-end training where the relative weighting of \textit{Stage~1} (V/A) and \textit{Stage~2} (questionnaire prediction; Q) losses shifts over the course of training. The losses are combined as a convex weighted sum, and training begins with $\lambda_\text{V/A} = 1.0$, $\lambda_\text{Q} = 0.0$, gradually transitioning to $\lambda_\text{V/A} = 0.3$, $\lambda_\text{Q} = 0.7$. This leverages the insight from \citet{terhurne2022validation} that intermediate affect representations are informative for questionnaire prediction, while progressively shifting the optimization focus toward the clinical targets.
\end{itemize}

\subsubsection{Evaluation protocol}
\label{sec:eval}

All experiments use 5-fold cross-validation with patient-level stratification, ensuring that all sessions from a given patient appear in the same fold. Within each training fold, an 85\%~/~15\% split reserves training patients for validation (early stopping); the model is then retrained with the selected epoch count on the full training fold. Both the deep learning and classical ML pipelines use identical fold assignments generated from a shared split definition, enabling direct comparison of results.

Performance is reported as Pearson correlation $r$ between predicted and observed values, averaged across folds. For frame-level V/A prediction, correlations are computed per session and averaged. For session-level questionnaire prediction, correlations are computed across all test sessions within each fold. We report mean $\pm$ standard deviation across the five folds.

\subsection{Results}

\subsubsection{Deep Learning vs.\ Classical ML Baselines}
\label{sec:baseline_results}

Table~\ref{tab:va_results_raw} summarizes the performance of all encoder--aggregation configurations on both frame-level V/A prediction and session-level questionnaire prediction under 5-fold patient-stratified cross-validation.

\textbf{Frame-level affect prediction.}
Frame-level V/A prediction reveals a clear hierarchy among encoder families. LSTM-based encoders performed poorly across all aggregation strategies, with average V/A correlations ranging from $.038$ (LSTM + Attention) to $.084$ (LSTM + LSTM). Transformer encoders substantially outperformed LSTMs, achieving $.23$--$.25$ average V/A correlation regardless of aggregation strategy. Among classical ML baselines operating on per-frame scalar features, MLP reached the highest average V/A correlation ($.274 \pm .041$), slightly exceeding the best transformer; XGBoost ($.269$) and Random Forest ($.254$) fell at or above the transformer range, while Ridge ($.216$) performed below it. This pattern suggests that for frame-level affect prediction from seven interpretable scalar features, temporal modeling within 2-second windows provides limited benefit over independent per-frame prediction.

\textbf{Session-level questionnaire prediction.}
At the session level, questionnaire prediction results differ between single-configuration averages and best-per-scale comparisons. Across all encoder--aggregation configurations (Table~\ref{tab:va_results_raw}), classical ML achieves a best average questionnaire correlation of $.071$ (Random Forest) against $.078$ for the best DL configuration (Transformer + Statistical pooling). When the best model is selected per questionnaire scale (Table~\ref{tab:quest_results_raw}), DL achieves a higher average ($.122$) than classical ML ($.084$), winning 5 of 9 scales. The DL advantage is concentrated on the well-being scales, two in-session mood ratings, and the anxiety subscale (EPO-1\textsubscript{Likert} = $.219$, EPO-1\textsubscript{VAS} = $.229$, POMS\textsubscript{Pat,Neg} = $.098$, POMS\textsubscript{Ther,Pos} = $.164$, HSCL-11\textsubscript{Anxiety} = $.053$), while classical ML leads on the remaining symptom scales (HSCL-11\textsubscript{Depression} = $.184$, HSCL-11\textsubscript{Overall} = $.094$) and the remaining mood ratings (POMS\textsubscript{Pat,Pos} = $.242$, POMS\textsubscript{Ther,Neg} = $.114$).

\begin{table*}[t]
\centering
\caption{Prediction performance (Pearson $r$, 5-fold patient-stratified CV, mean $\pm$ sd). Best classical ML result per column in \textbf{bold}; best DL result \underline{underlined}.}
\label{tab:va_results_raw}
\begin{tabular}{llcccc}
\toprule
Encoder & Aggregation & Valence $r$ & Arousal $r$ & Avg V/A $r$ & Avg Quest.\ $r$ \\
\midrule
\multicolumn{6}{l}{\textit{Classical ML (frame-level, scalar features)\textsuperscript{$\dagger$}}} \\
Ridge           & --          & $\phantom{-}.221 \pm .083$ & $\phantom{-}.212 \pm .034$ & $\phantom{-}.216 \pm .028$ & $\phantom{-}.069 \pm .175$ \\
XGBoost         & --          & $\phantom{-}.251 \pm .077$ & $\phantom{-}\mathbf{.288 \pm .073}$ & $\phantom{-}.269 \pm .040$ & $\phantom{-}.056 \pm .173$ \\
Random Forest   & --          & $\phantom{-}.241 \pm .076$ & $\phantom{-}.268 \pm .075$ & $\phantom{-}.254 \pm .043$ & $\phantom{-}\mathbf{.071 \pm .177}$ \\
MLP             & --          & $\phantom{-}.261 \pm .080$ & $\phantom{-}.286 \pm .077$ & $\phantom{-}\mathbf{.274 \pm .041}$ & $\phantom{-}.049 \pm .171$ \\
\midrule
\multicolumn{6}{l}{\textit{Deep Learning (window-level, sequence length = 50)}} \\
LSTM        & Statistical & $\phantom{-}.097 \pm .093$ & $\phantom{-}.042 \pm .083$ & $\phantom{-}.070 \pm .084$ & $\phantom{-}.014 \pm .060$ \\
LSTM        & Attention   & $\phantom{-}.075 \pm .086$ & $\phantom{-}.002 \pm .044$ & $\phantom{-}.038 \pm .029$ & $-.050 \pm .045$ \\
LSTM        & LSTM        & $\phantom{-}.113 \pm .090$ & $\phantom{-}.056 \pm .035$ & $\phantom{-}.084 \pm .056$ & $\phantom{-}.012 \pm .096$ \\
Transformer & Statistical & $\phantom{-}.279 \pm .065$ & $\phantom{-}.184 \pm .111$ & $\phantom{-}.231 \pm .052$ & $\phantom{-}\underline{.078 \pm .043}$ \\
Transformer & Attention   & $\underline{\phantom{-}.295 \pm .064}$ & $\underline{\phantom{-}.208 \pm .043}$ & $\underline{\phantom{-}.251 \pm .033}$ & $\phantom{-}.076 \pm .044$ \\
Transformer & LSTM        & $\phantom{-}.260 \pm .094$ & $\phantom{-}.212 \pm .080$ & $\phantom{-}.236 \pm .055$ & $\phantom{-}.034 \pm .064$ \\
\bottomrule
\end{tabular}

\smallskip
\noindent\textsuperscript{$\dagger$}Classical ML: V/A via per-frame prediction (7 scalar features); questionnaires via Ridge on session-aggregated features + V/A predictions from each model. Same patient-stratified folds as DL.
\end{table*}

\begin{table}[t]
\centering
\caption{Per-scale session-level questionnaire prediction (Pearson $r$, 5-fold patient-stratified CV). Best classical ML and best DL model selected per scale. Best per scale in \textbf{bold}.}
\label{tab:quest_results_raw}
\begin{tabular}{lcc}
\toprule
Questionnaire & Best Classical ML & Best DL \\
\midrule
\multicolumn{3}{l}{\textit{HSCL-11 (symptom burden)}} \\
Overall    & $\phantom{-}\mathbf{.094 \pm .244}$ & $\phantom{-}.058 \pm .216$ \\
Depression & $\phantom{-}\mathbf{.184 \pm .168}$ & $-.001 \pm .122$ \\
Anxiety    & $-.077 \pm .303$ & $\phantom{-}\mathbf{.053 \pm .115}$ \\
\midrule
\multicolumn{3}{l}{\textit{EPO-1 (well-being)}} \\
Likert & $\phantom{-}.121 \pm .207$ & $\phantom{-}\mathbf{.219 \pm .156}$ \\
Visual Analog Scale & $\phantom{-}.167 \pm .227$ & $\phantom{-}\mathbf{.229 \pm .174}$ \\
\midrule
\multicolumn{3}{l}{\textit{POMS (mood states)}} \\
Pat.\ Neg.  & $-.039 \pm .159$ & $\phantom{-}\mathbf{.098 \pm .154}$ \\
Pat.\ Pos.  & $\phantom{-}\mathbf{.242 \pm .244}$ & $\phantom{-}.199 \pm .206$ \\
Ther.\ Neg. & $\phantom{-}\mathbf{.114 \pm .245}$ & $\phantom{-}.082 \pm .245$ \\
Ther.\ Pos. & $-.047 \pm .121$ & $\phantom{-}\mathbf{.164 \pm .173}$ \\
\midrule
\textit{Average} & $\phantom{-}.084$ & $\phantom{-}\mathbf{.122}$ \\
\bottomrule

\end{tabular}
\end{table}

\subsection{Discussion}
The dissociation between frame-level V/A prediction and session-level questionnaire prediction in the deep learning pipeline is informative. Transformer encoders learn meaningful frame-level affect representations as average V/A correlations of $.23$--$.25$ represent a substantial signal, yet this signal fails to propagate into consistent session-level predictions through any of the learned aggregation strategies (Table~\ref{tab:va_results_raw}). The most likely explanation is data scarcity at the session level: with approximately 70 training sessions per fold (after reserving 15\% for validation), the aggregation modules and prediction heads lack sufficient examples to learn the mapping from continuous affect trajectories to clinical outcomes. This bottleneck is likely compounded by substantial between-patient heterogeneity and within-patient variability, which may make the affect–outcome mapping highly patient- and session-dependent. Statistical pooling mitigates this problem by replacing learned aggregation with fixed summary statistics, thereby converting the task into tabular regression with interpretable features, a regime in which classical ML is known to perform well~\cite{grinsztajn2022tree}.

At the questionnaire level, DL shows a slight overall advantage (Table~\ref{tab:quest_results_raw}), with its edge concentrated on the well-being and mood scales while classical ML retains an advantage on the symptom scales. However, the two-stage DL pipeline requires intermediate V/A supervision and multi-stage training, substantially increasing complexity and computational cost relative to the classical ML baseline.

Given this trade-off, we pursue a two-part investigation. Study~2 conducts a systematic ablation of the classical ML pipeline (varying prediction model, feature smoothing, session segmentation, and modality selection) to identify the optimal statistical aggregation configuration. Study~3 then revisits deep learning with a simplified single-stage architecture, incorporating findings from Study~2 for a cost-comparable re-evaluation.

\section{Study 2: Statistical Pooling of Frozen Embeddings with Classical ML}

Study 2 asks which representation and preprocessing choices are most effective. Starting from statistical pooling of the frozen pretrained embeddings, it systematically ablates prediction models, feature smoothing, session slicing, modality subsets, therapist features, and the merit of continuous human annotations.

\subsection{Method}
\label{sec:ablation}

Building on the baseline observations that learned aggregation failed to translate frame-level signal into consistent session-level predictions and that the best deep configuration itself relied on statistical pooling, we conducted a systematic ablation study to optimize performance across the design space. The study proceeds in five steps, each narrowing the configuration space based on the preceding results. All ablations use the same 5-fold patient-stratified cross-validation protocol described in Section~\ref{sec:eval}.
All input features are z-scored with dataset-level statistics computed on the training folds only, refit within each cross-validation split.
To favor fold-to-fold consistency over the raw correlation value, we rank configurations by the lower bound of the 95\% confidence interval computed as   \[
\mathrm{CI}_{95} =
\left[
\mathrm{corr} - t_{0.975,4}\,\mathrm{SEM},
\mathrm{corr} + t_{0.975,4}\,\mathrm{SEM}\right]
\]
using the standard error of the mean (SEM) and the Student's \textit{t}-distribution with  $\mathit{n}-1$ degrees of freedom for the number of folds $n=5$.

The five ablation steps systematically test the performance of each classical ML model from Section~\ref{sec:stage2}, the influence of feature smoothing, and the effective session window, across all patient feature combinations. We then assess the change in performance when therapist features are added to the best model--feature combinations, and finally analyze the impact of single modalities and the merit of the human annotations described in Section~\ref{sec:humananno}.

\subsection{Results}

\subsubsection{Model Ablation}

\begin{figure*}[!t]
    \centering
    \includegraphics[width=\linewidth]{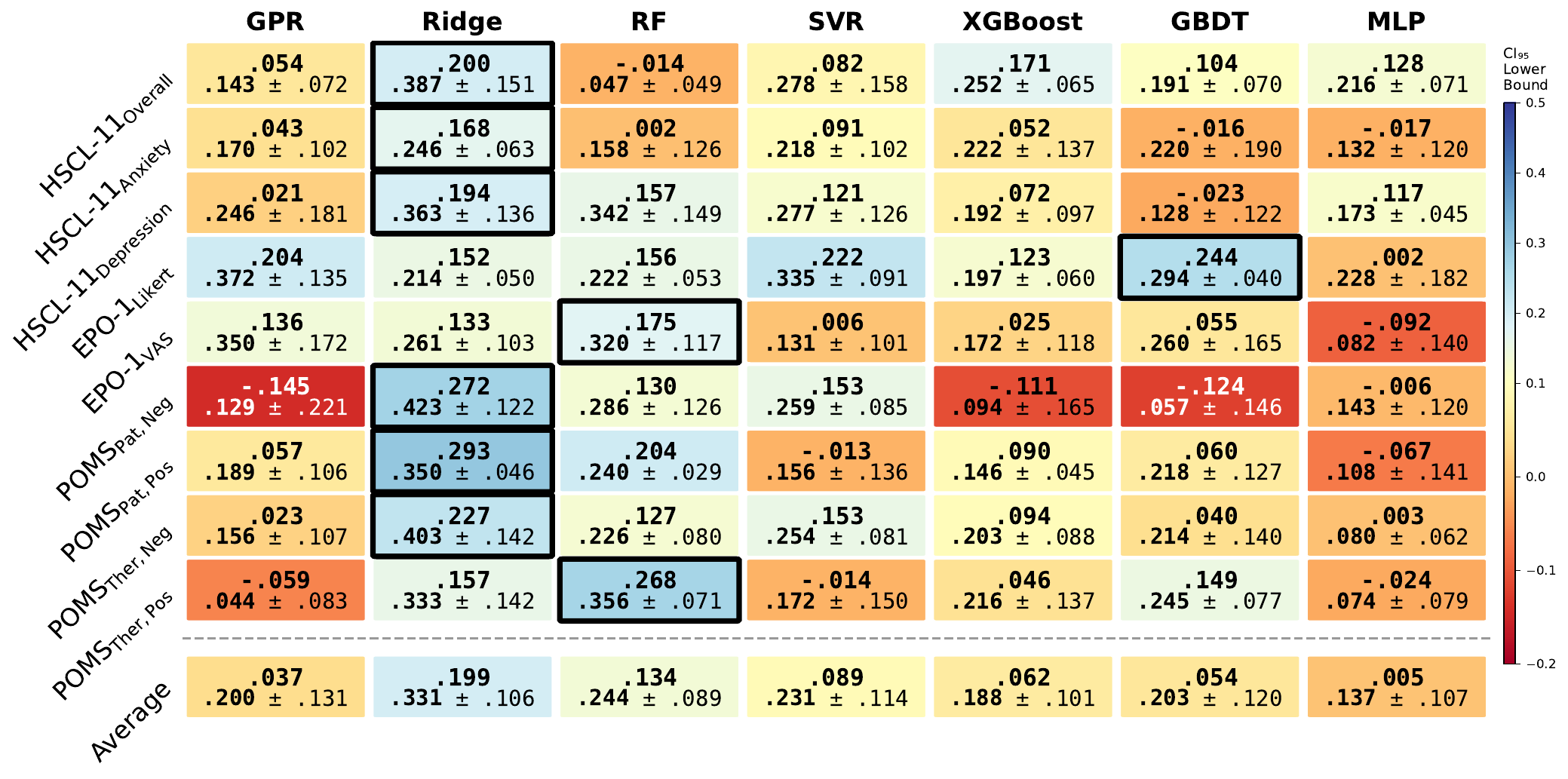}
    \caption{\textbf{Results of the model ablation study.} Rows correspond to questionnaires and subscales, and columns to predictive models. Each cell reports the lower bound of the CI$_{95}$ in the first line and the mean correlation ± standard deviation in the second line for the best performing input features. Cell shading encodes the lower 95\% confidence bound on a diverging scale. Black outlines mark the best-performing cells according to this criterion. No human features were used and no smoothing was applied.}
    \label{fig:abl_1}
\end{figure*}

Our first goal was to compare all feature combinations defined in Section \ref{sec:featureextrac} for the nine questionnaire-derived outcome variables (See Section \ref{sec:Questionaires}): three HSCL-11 scores (overall score and the anxiety and depression subscales), two EPO-1 scores (Likert and VAS), and four POMS scores (positive and negative emotion, each rated separately by patients and therapists).

We predict them using seven different classical machine learning models suited to low-data regimes. To cover a broad range of approaches, we evaluated the linear regression model Ridge, the kernel-based models Gaussian Process Regression (GPR) and Support Vector Regression (SVR), the ensemble-based models Random Forest (RF), XGBoost, and Gradient Boosted Decision Trees (GBDT), and a basic neural network using a two-layer MLP with distinct heads for each questionnaire, a dropout rate of $0.2$, and hidden dimensions of 128 and 64 for the two layers respectively. For each feature dimension, we calculated the mean, standard deviation, minimum, maximum, and median of each session. Overall, 15,408 feature-model-questionnaire combinations were trained for this ablation step. The results of this ablation can be seen in Figure \ref{fig:abl_1}.

GPR, SVR, XGBoost, and MLP had low average prediction correlations. GBDT had the best output for a single questionnaire (EPO-1\textsubscript{Likert} $r=.294 \pm .040$) but showed subpar overall performance.
RF performed best on the EPO-1\textsubscript{VAS} questionnaire ($r=.320 \pm .117$) and POMS\textsubscript{Ther,Pos} questionnaire ($r=.356 \pm .071$). The best prediction quality was achieved by Ridge, which performed comparatively high overall, being the best model on average and for 6 out of the 9 questionnaires:
HSCL-11\textsubscript{Overall} ($r=.387 \pm .151$), HSCL-11\textsubscript{Anxiety} ($r=.246 \pm .063$), HSCL-11\textsubscript{Depression} ($r=.360 \pm .134$), both POMS\textsubscript{Pat} questionnaires (positive: $r=.350 \pm .046$, negative $r=.423 \pm .122$) and the POMS\textsubscript{Ther,Neg} questionnaire ($r=.403 \pm .142$).

Similar to the initial prediction performance in Table \ref{tab:va_results_raw}, Ridge and RF show the strongest overall results with average correlations of $r=.331 \pm .106$ and $r=.244 \pm .089$ respectively. RF performed best of all ensemble models and Ridge best of all regression models and overall. We continue with the two strongest models that perform best on eight of the nine questionnaires: Ridge and RF.

\subsubsection{Smoothing Ablation}

The aim of our second ablation step was to analyze feature smoothing and its effect on model performance. The handcrafted output features and final modal embeddings are calculated on a frame-wise level and are often subject to unforeseen noise, inconsistencies and erroneous data output for small window sizes that influence the session minima and maxima~\cite{PORIA201798}. Additionally, averaging over the entire session from a frame-wise output could miss overall and longer-term behavioral patterns that emerge only event-based~\cite{Lingenfelser2018} or over longer periods of time~\cite{fernando2016rank}.

To address these possible issues, we decided to systematically investigate the effect of feature smoothing using 15 smoothing settings: 14 window sizes plus the entire session. We smooth the features using a sliding causal window with no zero-padding at the session start. Window sizes follow a geometric (doubling) progression anchored at a one-second (25-frame) window and extended in both directions, ranging from a single frame (no smoothing) up to 25{,}600 frames ($\approx$17 minutes), plus smoothing over the entire session ($\approx$50 minutes on average). A full ablation of all feature and embedding combinations for each questionnaire and window size was run over the four retained models (Ridge, RF, GBDT, and XGBoost; 147,753 trained combinations); Fig.~S1 in the supplementary material displays the Ridge and RF results.

The strongest-performing smoothing window on average was $w=200$ frames with $r=.343 \pm .089$, closely followed by no smoothing ($r=.341 \pm .100$; this average pools the best of Ridge and RF per scale and therefore sits slightly above the Ridge-only average of $.331$ in Figure~\ref{fig:abl_1}). Other well-performing window sizes were 100 ($r=.336 \pm .114$), 400 ($r=.301 \pm .100$), 12800 ($r=.325 \pm .114$) and 25600 ($r=.319 \pm .108$).

The results indicate that the ideal smoothing size is dependent on the questionnaire. Four window sizes showed particularly high correlations for at least one questionnaire, reflecting distinct peaks for no, light, medium, and heavy smoothing: no smoothing ($w=1$; POMS\textsubscript{Pat,Neg} $r=.423 \pm .122$, POMS\textsubscript{Pat,Pos} $.350 \pm .046$, POMS\textsubscript{Ther,Neg} $.403 \pm .142$), light ($w=200$; HSCL-11\textsubscript{Anxiety} $.312 \pm .112$, EPO-1\textsubscript{VAS} $.271 \pm .053$, POMS\textsubscript{Ther,Pos} $.456 \pm .078$), medium ($w=1600$; HSCL-11\textsubscript{Depression} $.493 \pm .033$), and heavy ($w=25600$; HSCL-11\textsubscript{Overall} $.426 \pm .075$, EPO-1\textsubscript{Likert} $.337 \pm .091$). We therefore use these four window sizes for the further ablation.

\subsubsection{Session Slicing Ablation}

The third goal was to analyze the predictive value of different session slices. We chose four parts of the session: the first, middle, and final third, and the \emph{working phase}, i.e.\ the 15 minutes preceding the final 5 minutes of the session, considered the period in which patients are most likely to engage in deep therapeutic work~\cite{auszra2013client}. We also evaluated the three thirds combined.

Again, we tested all feature input combinations for all nine questionnaires, four window sizes, and the four retained models for a total of 217,332 trained models. Fig.~S2 in the supplementary material shows the Ridge and RF results; the per-scale winning configurations, including those from the boosted models, are summarized in Fig.~S3 in the supplementary material.

Slicing improved the CI$_{95}$-based selection score on five of the nine questionnaires. Using only the middle of each session improved the EPO-1\textsubscript{VAS} ($.271 \pm .053 \to .404 \pm .074$), using the beginning of each session improved the consistency of the POMS\textsubscript{Pat,Neg} ($.423 \pm .122 \to .366 \pm .076$), the working phase improved the POMS\textsubscript{Pat,Pos} ($.350 \pm .046 \to .447 \pm .088$) and HSCL-11\textsubscript{Anxiety} ($.312 \pm .112 \to .262 \pm .061$), and the combination of early, mid, and late performed best on HSCL-11\textsubscript{Overall} ($.426 \pm .075 \to .490 \pm .053$). For the therapist-feature step, we use the best three performing model, window and slicing combinations for each questionnaire to simplify the therapist validation.

\subsubsection{Therapist Feature Ablation}

Psychotherapy is an inherently dyadic process. Therapeutic process research links in-session emotional expression and the patient--therapist interaction to treatment outcome~\cite{peluso2018therapist}. We therefore hypothesize that information about therapist behavior can improve questionnaire prediction performance. The fourth ablation step therefore adds therapist features for all modalities to the best performing model--feature--window combinations. We used the three best settings for each questionnaire from the preceding ablation steps and append the same handcrafted features and embeddings of the therapist to the model input. Crossing the full configuration grid with therapist-feature variants would be combinatorially infeasible, and this choice follows the sequential-ablation logic used in the preceding steps, where each step builds on the retained best configurations from the previous one. This lets the model condition on therapist behavior and displayed emotion alongside the patient features, which could improve prediction. To test how the therapist features perform independently, we also trained the models purely on therapist features. The results of these additions as well as the final best model results can be seen in Fig.~S4 in the supplementary material.

Patient-side features alone proved sufficient: adding therapist features did not improve the CI$_{95}$-based ranking on any questionnaire, whether combined with patient features or used alone, and therapist-only models performed markedly worse throughout (average $r = .406$ patient-only, $.336$ patient+therapist, $.285$ therapist-only; Fig.~S4). For deployment this is a welcome result, as a single patient-facing capture stream carried the outcome-relevant signal in our feature space, reducing recording and preprocessing burden and simplifying consent, without sacrificing performance.
Notably, seven of the nine outcome scales are patient-reported, while the two therapist-rated POMS scales likewise refer to the patient’s state and
were predicted best from patient features alone as well (e.g.\ POMS\textsubscript{Ther,Pos}: $.456$ patient-only vs.\ $.406$ when therapist features are added), indicating that patient-side features provide the more direct predictive signal for these targets.
As all features enter the models as per-person summaries without dyadic interaction terms, this analysis cannot speak to synchrony- or alliance-based accounts of therapist influence.

\subsubsection{Modality Comparison and Human Annotations}

\begin{figure*}[!t]
    \centering
    \includegraphics[width=\linewidth]{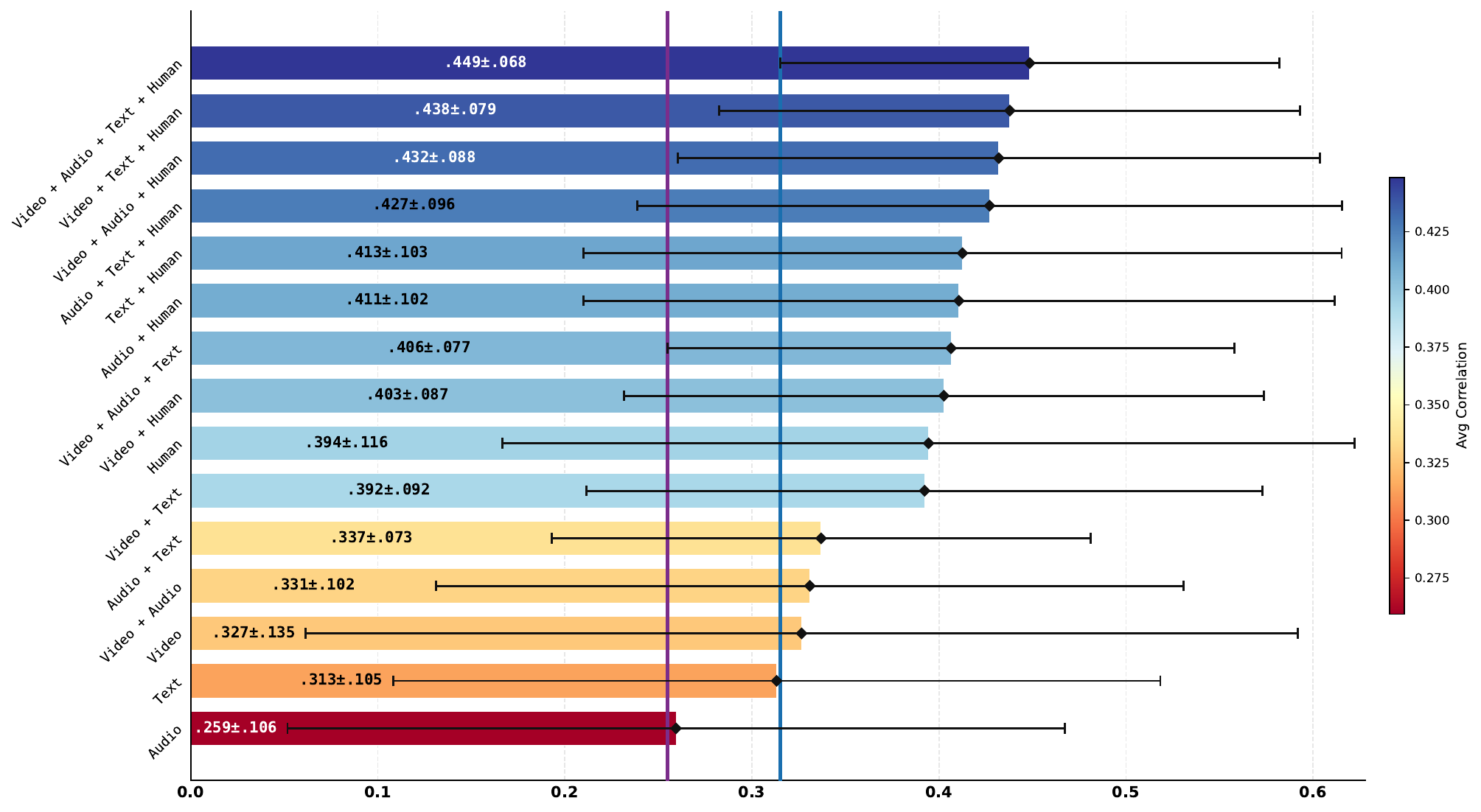}
    \caption{\textbf{Comparison of best questionnaire performance of all modalities.}
    Video represents EmoNet features, audio represents eGeMAPS and EmoW2V features, text represents XLM-T sentiment features, and Human represents the manually annotated V/A values. Human annotations are available only for these 106 sessions. Whiskers are CI\textsubscript{95} bounds.
    Blue line: lower bound for all features; purple: lower bound for all features except human annotations.}
    \label{fig:abl_5}
\end{figure*}

In this subsection, we compare the performance of each available input feature and their combinations. Here, we only focus on overall average performance over all 9 questionnaires analyzed in the previous ablations. A visual representation can be seen in Figure \ref{fig:abl_5}.

While it is essential to know which multimodal model inputs produce the best possible prediction results, some modalities may not be available or easily accessible in routine clinical settings. For example, many practices may not be able to implement video recordings of therapy sessions, whereas audio recordings and automatically derived text transcripts may be more feasible. Therefore, it is important to understand the predictive value of individual modalities and their combinations in order to assess the applicability of questionnaire prediction under different clinical constraints.

Additionally, we have also manually annotated V/A values for the initial dataset of 106 sessions as introduced in Section \ref{sec:humananno}. These annotations are very laborious to produce and difficult to scale. To investigate whether the annotation effort yields a meaningful predictive benefit, we compare model output with and without those annotations. The direct feature comparison of model performance including human annotations and excluding human annotations (including subsets of features) can be found in Fig.~S3.

Text is the strongest single modality, already reaching $r = .313 \pm .105$ on its own (ranked above video by the CI$_{95}$ lower bound), yet every modality combination outperforms every single modality, underscoring their complementarity. Video and text together perform best among the two-modality sets at $r = .392 \pm .092$; the fully camera-free audio-plus-text combination reaches $r = .337 \pm .073$. Subsets of all three extracted modalities increase performance to $r = .406 \pm .077$. Figure~\ref{fig:abl_5} ranks all modality combinations by their CI$_{95}$ lower bound.

These increments map directly onto the recording constraints of routine practice, where the available channels vary from site to site. Text, obtainable from any audio recording via automatic transcription, is the backbone: it is the strongest single modality ($r = .313$) and anchors every strong combination. A practice that keeps only transcripts and discards audio therefore retains most of the usable single-stream signal, while a site that records audio but no video reaches the fully camera-free audio-plus-text operating point ($r = .337$). Where a camera is acceptable, adding video yields the largest increment (video-plus-text, $r = .392$) and recovers most of the full three-modality performance ($r = .406$), so a clinic that can capture video gains more from it than from audio. In short, text sets a usable floor that survives the most restrictive setups, and each additional modality buys a modest, diminishing increment rather than a step change, which lets a deployment trade recording burden against expected accuracy according to what its setting permits.

Adding human V/A annotations raises average performance to $r = .449 \pm .068$, the highest average of any single feature set (with per-scale model, window, and segment choices as in the preceding ablations).
Human annotations alone, however, reach only $r = .394 \pm .116$, below the combined video, audio, and text modalities.
Across the nine scales, the paired difference between including and excluding human annotations is small and does not reach significance
(per-scale best configurations: averages $.448$ with vs.\ $.406$ without human annotations; improvements on seven of nine scales, decreases on two, median $\Delta = +.026$; Wilcoxon signed-rank $W = 8$, $p = .098$, $n = 9$).
The automatic features thus already capture the session-level signal contained in the continuous annotations, indicating that the pipeline can operate fully annotation-free.

A per-scale comparison (Fig.~S3) ranks each scale by the CI$_{95}$ lower bound: the human-inclusive set ranks higher on eight of nine scales (the exception is EPO-1\textsubscript{VAS}, $.402$ vs.\ $.404$). On POMS\textsubscript{Ther,Neg} it ranks higher despite a lower mean ($.314 \pm .045$ vs.\ $.403 \pm .142$), because the variance-penalizing criterion rewards its lower fold spread. The largest gap is on POMS\textsubscript{Pat,Neg} ($r = .539 \pm .027$ with vs.\ $.366 \pm .076$ without). Per-scale fold-level tests ($n = 5$) are underpowered by construction, so we read these differences as descriptive. Given the real-time annotation effort ($\approx$50 minutes per session), automatic extraction attains statistically indistinguishable performance at negligible marginal cost, removing the principal scalability bottleneck of annotation-based pipelines.

\subsection{Discussion}
The ablation establishes statistical pooling of pretrained embeddings with regularized classical models as a strong session-level predictor, reaching $r = .406$ across nine scales under per-scale configuration selection ($r = .406$ for the best fixed three-modality feature set) and $r = .490$ for overall symptom distress. Regarding the choice of feature combinations, the winning configurations draw on the pretrained embeddings rather than the models' scalar affect outputs alone: six of the nine per-scale best feature sets include at least one embedding stream (Fig.~S3), and the best configuration for the shared HSCL-11 overall score uses embeddings exclusively. The modalities are moreover complementary: no single modality approaches the combined feature sets (text, the strongest, reaches only $r = .313$ alone), while video-plus-text already recovers most of the full multimodal performance ($.392$ of $.406$). Importantly, the ablations show that performance can be further improved through targeted per-scale configuration choices. Session slicing improved several questionnaires, with particularly pronounced gains for symptom distress and well-being (HSCL-11\textsubscript{Overall} $.426 \to .490$, EPO-1\textsubscript{VAS} $.271 \to .404$), suggesting that clinically relevant predictive information is not uniformly distributed across the session but may depend on the temporal segment considered. Feature smoothing and slicing also contributed to more stable estimates in some settings by reducing fold-to-fold variance. In contrast, therapist features did not add reliable predictive value in the present setup. This may partly reflect the fact that most target questionnaires were based on patient self-reports, making patient-side embeddings more directly aligned with the outcomes than therapist-side features. Finally, although the single highest-scoring feature set included the labor-intensive human valence/arousal annotations, it did not significantly outperform fully automatic features ($p = .098$). Together, these findings suggest that automatic patient-side embeddings already capture a substantial share of the predictive signal, while carefully selected, questionnaire-specific configuration choices (feature subset, model, smoothing window, and segment) provide a practical and low-cost route to improved performance. This motivates Study 3, which asks whether a single learned model can match this pooled-ML pipeline without requiring per-scale configuration selection.

\section{Study 3: Single-Stage Deep Learning Incorporating Study-2 Lessons}
\label{sec:dl_revisited}

Study 3 asks whether deep learning remains inferior once the representation and preprocessing choices optimized in Study~2 are held to their strongest observed form. It replaces the two-stage curriculum with a simplified single-stage architecture trained directly on questionnaire targets.

\subsection{Method}

The baseline deep learning pipeline (Section~\ref{sec:baseline_results}) achieved a slight average advantage over classical ML for session-level questionnaire prediction, but required a multi-stage training curriculum with intermediate V/A supervision, making it substantially more complex and costly to train. We therefore revisit deep learning with a simplified, single-stage architecture trained end-to-end directly on questionnaire targets. Its design adopts the train-fold z-scoring regime introduced with the ML ablation and builds on Study~1's observation that short-term ($\leq$2\,s) windows already carry predictive affective signal.

\subsubsection{Architecture}

The revised pipeline replaces the frame-level V/A encoder and multi-stage curriculum with a single session-level model:

\begin{enumerate}
    \item \textbf{Feature extraction (frozen).} Per-frame features from three pretrained models: EmoNet (256-d embedding + V/A scalars), EmoW2V (1024-d embedding + V/A/dominance scalars), XLM-T (768-d embedding + sentiment scalar). The eGeMAPS descriptors were omitted, as none of the best-performing ML configurations relied on them. Features are z-scored using train-fold statistics only, identical to the ML ablation protocol.
    \item \textbf{Windowing.} Sessions are segmented into overlapping windows of $W=50$ frames ($\approx$2\,s at 25\,fps) with stride $W/2$ (50\% overlap). Each window is summarized by mean pooling.
    \item \textbf{Session aggregation.} An attention-based aggregator learns a weighted combination of window representations (hidden dim $= 64$, tanh-MLP scoring).
    \item \textbf{Multi-task prediction.} Nine parallel MLP heads (Linear $\to$ ReLU $\to$ Dropout $\to$ Linear, hidden dim $= 32$) predict the nine questionnaire scores jointly under a multi-task MSE loss.
\end{enumerate}

Training uses Adam (lr\,$=10^{-3}$, wd\,$=10^{-4}$), 15 epochs, same 5-fold patient-stratified CV as the ML ablation. The best epoch is selected by validation loss (Phase~1); the final model is retrained on train\,+\,val for that epoch count (Phase~2).

\subsubsection{Configuration selection}

We conducted a systematic ablation over the architectural degrees of freedom before finalizing the model above. Window size $W \in \{50, 200, 800, 1600\}$ was swept on the attention-embedding cell; $W=50$ outperformed all others within the sweep ($r=.225$ vs.\ $.184$ at $W=200$); the final model retrained at $W=50$ reaches the reported $.222$. Aggregator $\times$ feature type (4\,$\times$\,2 cells) were evaluated at $W=50$; attention with pretrained embeddings (attention\_emb) achieved the highest average questionnaire correlation. Hidden dimension $d \in \{32, 64, 128, 256\}$ was swept; $d=64$ was the consistent winner. Per-questionnaire single-output training did not improve over the multi-task baseline (Wilcoxon $p=.910$, $n=9$), and a learnable per-frame projection left performance essentially unchanged ($.231$ vs.\ $.222$). Homoscedastic (Kendall-style) uncertainty weighting hurt performance ($.185$ vs.\ $.222$), consistent with the data-scarce regime.

\subsection{Results}

\subsubsection{Performance}
\label{sec:dl_results}

Table~\ref{tab:dl_quest_results} reports per-scale questionnaire prediction for the final single-stage DL model alongside the ML ablation oracle (best no-human configuration per scale; same folds).

\begin{table}[!t]
\centering
\caption{Per-scale session-level questionnaire prediction (Pearson $r$, 5-fold patient-stratified CV, mean\,$\pm$\,sd). \textbf{ML peak}: best no-human configuration per scale from the Study~2 ablation (ranked by the CI$_{95}$ lower bound). \textbf{DL}: single-stage attention model, pretrained embeddings, $W=50$. Best per scale in \textbf{bold}. The ML peak selects smoothing window and segment per scale; selection-tier comparisons are discussed in Section~\ref{sec:like_for_like}.}
\label{tab:dl_quest_results}
\begin{tabular}{lcc}
\toprule
Questionnaire & ML peak (no H) & DL (attn-emb, $W=50$) \\
\midrule
\multicolumn{3}{l}{\textit{HSCL-11 (symptom burden)}} \\
Overall        & $\phantom{-}\mathbf{.490 \pm .053}$ & $\phantom{-}.219 \pm .169$ \\
Depression     & $\phantom{-}\mathbf{.493 \pm .033}$ & $\phantom{-}.237 \pm .182$ \\
Anxiety        & $\phantom{-}\mathbf{.262 \pm .061}$ & $\phantom{-}.088 \pm .289$ \\
\midrule
\multicolumn{3}{l}{\textit{EPO-1 (well-being)}} \\
Likert & $\phantom{-}\mathbf{.337 \pm .091}$ & $\phantom{-}.223 \pm .093$ \\
Visual Analog Scale & $\phantom{-}\mathbf{.404 \pm .074}$ & $\phantom{-}.354 \pm .179$ \\
\midrule
\multicolumn{3}{l}{\textit{POMS (mood states)}} \\
Pat.\ Neg. & $\phantom{-}\mathbf{.366 \pm .076}$ & $\phantom{-}.144 \pm .174$ \\
Pat.\ Pos. & $\phantom{-}\mathbf{.447 \pm .088}$ & $\phantom{-}.309 \pm .185$ \\
Ther.\ Neg. & $\phantom{-}\mathbf{.403 \pm .142}$ & $\phantom{-}.191 \pm .095$ \\
Ther.\ Pos. & $\phantom{-}\mathbf{.456 \pm .078}$ & $\phantom{-}.233 \pm .158$ \\
\midrule
\textit{Average} & $\phantom{-}\mathbf{.406}$ & $\phantom{-}.222$ \\
\bottomrule
\end{tabular}
\end{table}

The single-stage DL model reaches an average questionnaire correlation of $.222$, a substantial improvement over the two-stage baseline ($.078$ best, Table~\ref{tab:va_results_raw}). The optimized ML ablation achieves $.406$ on average, with ML winning on all nine questionnaires. DL performance is uneven across folds (per-fold averages $.366$, $.243$, $.186$, $.121$, $.194$), consistent with a variance-dominated small-sample regime. These two figures sit at different selection tiers, however; the like-for-like comparison below locates where the ML advantage actually originates.

\subsubsection{Like-for-like model selection}
\label{sec:like_for_like}

The headline ML figure ($.406$) reflects per-scale configuration selection over the ablation grid, whereas the single deployable DL configuration ($.222$) does not. Comparing matched selection tiers separates the model class from the selection protocol. At the single-fixed-configuration tier (one feature set, model family, smoothing window, and segment applied to all nine scales), ML averages $.281$, ahead of the single deployable DL model ($.222$). Per-scale selection over the ablation grid (feature subset, model family, smoothing window, and segment) then widens the gap substantially: it lifts ML from $.281$ to $.406$, while granting DL the analogous per-scale oracle (best of eight aggregator/feature cells per scale) raises it only to $.252$, below ML on all nine scales. Most of the ML advantage thus stems from the inexpensive grid search that its low training cost makes feasible, on top of a modest single-configuration edge.

\subsubsection{Statistical comparison}
\label{sec:dl_stats}

We compare ML and DL using a pre-specified family of two confirmatory contrasts (ML vs.\ DL; with vs.\ without human annotations), with Holm--Bonferroni correction applied across that family; reported $p$-values are Holm-adjusted. For the headline ML vs.\ DL comparison, we treat the nine questionnaires as exchangeable replicates of the same scientific claim (predictability from affective dynamics), yielding a Wilcoxon signed-rank test on $n=9$ paired mean correlations (differences computed as DL minus ML). The result confirms the ML advantage: $W = 0$, Holm-adjusted $p = .008$, median difference $= -.212$ (ML wins all nine scales). Under the more conservative like-for-like contrast (per-scale oracle on both sides), the advantage persists ($W = 0$, $p = .004$, median difference $= -.162$, ML superior on all nine scales); this oracle contrast is reported unadjusted, as a sensitivity analysis outside the confirmatory family.

The DL fold-level standard deviations are far larger (mean $\sigma = .169$ vs.\ $.082$ for ML), consistent with the variance-dominated regime discussed below.

\subsection{Discussion}

The revised single-stage DL architecture substantially improves over the two-stage baseline, moving from near-chance performance ($.078$) to $.222$, a near-threefold gain that validates the design lessons imported from Study~2. The remaining gap to the ML ablation peak ($.406$) is best interpreted as a structural consequence of dataset size rather than an architectural limitation. The effective training set ($n \approx 70$ sessions per fold) is sufficient for a fixed aggregation strategy (ML: statistical pooling onto tabular features) but underpowers a jointly learned session representation. Consistent with this, the four architectural improvements tested (larger windows, learned projection, per-questionnaire heads, uncertainty weighting) either hurt or made no difference, a signature of the variance-dominated regime.

The finding that classical ML with statistical pooling outperforms learned deep aggregation under data scarcity is in line with prior benchmarks~\cite{grinsztajn2022tree}. Therefore it motivates the use of pretrained embeddings as frozen feature extractors rather than fine-tuned end-to-end components in small-$n$ affective computing settings.

\section{General Discussion}
\label{sec:gen-disc}

Across three studies, a consistent picture emerges of how pretrained multimodal representations are best leveraged for clinical outcome prediction under data scarcity. The two-stage pipeline with intermediate V/A supervision (Study~1) learned meaningful frame-level affect representations but transferred little of that signal to the session level ($r = .078$ for the best single configuration; $.122$ under per-scale selection). Replacing learned aggregation with statistical pooling and classical prediction models, then systematically optimizing that pipeline (Study~2), raised average performance to $r = .406$ across nine clinical scales. A simplified single-stage deep architecture incorporating the lessons of Study~2 (Study~3) reached $r = .222$ as a single deployable configuration, below the analogous single-configuration ML model ($.281$), and remained below the per-scale ML ablation, which is superior on all nine scales under matched per-scale selection. The decisive factor is not representational capacity but estimation variance: with roughly 70 training sessions per fold, fixed summary statistics feeding regularized models extract the available signal more reliably than any learned aggregation we tested.

Notably, the pooled classical pipeline reaches an average correlation of $.406$ under per-scale selection, well above both the single-configuration deep model ($.222$) and its per-scale oracle ($.252$). The limiting factor may, however, not have been the modalities themselves, but how information from these modalities was represented and aggregated. The shift from scalar affect outputs to pretrained embeddings with richer aggregation, smoothing, and session slicing allowed video-plus-text and full multimodal configurations to reach $r = .392$ and $.406$, indicating that clinically relevant signal is present in the recorded channels but can be lost when reduced to overly coarse features.

The absence of added value from therapist-side features may have several, mutually compatible explanations.
Seven of the nine outcome scales are patient-reported, while the two therapist-rated POMS scales likewise refer to the patient’s state. Also, the therapist-rated scales were predicted best from patient features alone in the therapist ablation, suggesting that the strongest available predictive signal for these patient-centered outcomes may be expressed in the patient stream. Consistent with this, associations between patient self-reports and therapist ratings are typically only moderate~\cite{lutz2021measuring}. At the same time, all features enter the models as per-person summary statistics without dyadic or interaction terms, so synchrony-, attunement-, or alliance-based mechanisms of therapist influence are structurally invisible to the models. Therapist behavior may also be affected by professional display regulation: therapists deliberately modulate their expressed affect, which could reduce between-session variance and thus discriminative signal in the therapist streams. Finally, with 53 therapists across 106 sessions, therapist-identity heterogeneity may add variance that roughly 70 training sessions per fold cannot average out. The added features may have introduced noise without capturing clinically meaningful process information. These accounts are testable with dyadic features and larger samples, and the null result should not be read as evidence that therapist behavior is clinically irrelevant.

The temporal findings point to a similar distinction between useful fixed structure and hard-to-estimate learned structure. The dimension of time is not entirely absent from the successful pipeline: session slicing is a coarse form of temporal selection and improved several scales, including symptom distress from $.426$ to $.490$ and well-being VAS from $.271$ to $.404$. This suggests that coarse temporal weighting may suffice for the present targets, whereas fine-grained sequence modeling adds little at this sample size. Session-level questionnaire outcomes are slow-moving, partly trait-like quantities, so the moment-to-moment ordering of affective states may genuinely carry little additional information for them, unlike tasks such as rupture detection where ordering is constitutive. This interpretation also matches the estimation-variance account from Study~3: learned temporal aggregation is precisely the component that roughly 70 sessions per fold cannot reliably fit, while fixed pooling spends no sample budget on learning how to aggregate.

\subsection{Practical implications}

Two results carry direct practical weight for scaling session-level measurement. First, continuous human V/A annotation, which requires trained raters, ongoing supervision, and approximately real-time effort per session hour, improved average prediction non-significantly ($r = .448$ vs.\ $.406$) with gains above $.1$ on only two of nine scales (POMS\textsubscript{Pat,Neg}, HSCL-11\textsubscript{Anxiety}). Second, therapist-side features produced no consistent improvement over patient-only input. For future routine outcome monitoring, the favorable operating point is therefore fully automated extraction from the patient stream alone: frozen pretrained extractors, statistical pooling, and a regularized prediction model, as a hypothesis-generating recommendation pending external validation. Because feature extraction is decoupled from prediction and all processing runs locally, this configuration also retains the privacy advantages discussed in the introduction. Where recording constraints permit only a subset of channels, the modality ablation suggests prioritizing text and video capture, which together recover most of the full multimodal performance; where video recording is infeasible, audio-plus-text still offers a usable camera-free operating point ($r = .337$). We report the configuration-level results of the ablation grid (per-scale model, window, and segment choices) as descriptive evidence to inform such deployments; a dedicated best-practice evaluation in routine care remains future work.

\section{Limitations}

Several limitations bound the scope of our conclusions. First, the development sample comprises 106 sessions from 88 patients at a single university outpatient clinic, with German-language sessions and an integrative CBT treatment model; generalization to other settings, languages, and treatment approaches remains to be demonstrated, and validation on a larger routine-care sample is the most important next step. Second, $n=88$ patients with 5-fold cross-validation limits statistical power. We therefore report fold-level paired comparisons on a small set of pre-specified scientific contrasts (ML vs.\ DL, with vs.\ without human annotations) with Holm--Bonferroni correction across those contrasts only. The full ablation grid serves as descriptive evidence and is reported descriptively via per-scale CI$_{95}$ heat maps (Fig.~\ref{fig:abl_1}, Figs.~S1--S4) rather than per-cell tests; we focus on effect sizes and consistent rank patterns rather than per-cell significance. Third, most patients contribute only a single session, which limits conclusions about within-patient dynamics: the present analyses cannot reliably separate stable between-patient heterogeneity from session-to-session variation within the same patient. Future work should therefore validate the approach in a longitudinal routine-care sample with multiple sessions per patient and analyze affect--outcome associations using multilevel models that explicitly distinguish patient-level baselines from within-patient change. Fourth, selecting the best configuration per scale, even when ranked by the lower bound of the confidence interval, carries a risk of optimistic bias that only external validation can fully resolve; in particular, part of the per-scale gains from smoothing and slicing reflect fold-to-fold variance reduction rewarded by the CI$_{95}$ criterion rather than added signal. Fifth, speaker diarization relied on an oracle-based mapping of clusters to speakers, which a deployed system would need to replace with a fully automatic procedure. Finally, our models quantify associations between in-session affective dynamics and questionnaire scores. They are not calibrated estimators of individual clinical status and are not intended to replace validated assessment instruments.

\section{Conclusion}
We asked how pretrained multimodal representations are best leveraged for predicting clinical outcomes from psychotherapy sessions when data are scarce. The answer, across a systematic comparison of deep and classical pipelines and a large-scale ablation, is consistent: frozen pretrained embeddings, summarized by simple statistics over multiple temporal windows and fed to regularized classical models, outperform every learned aggregation strategy we tested once inexpensive per-scale configuration selection is applied, reaching an average correlation of $.406$ across nine clinical scales and $.490$ for overall symptom distress. Continuous human affect annotations did not add significant predictive value, and a carefully simplified single-stage deep model, while far better than its two-stage predecessor, confirmed that frozen embeddings with statistical pooling remain the stronger and cheaper operating point at this sample size. Therapist-side features likewise did not improve prediction on any scale, a pattern consistent with the patient-centered nature of the
outcome measures (see Sec.~\ref{sec:gen-disc}).
This should not be interpreted as evidence that therapist behavior is clinically irrelevant, as therapist contributions may operate through dyadic processes such as synchrony, responsiveness, or alliance that are not captured by separate per-person summary statistics. Our results position richly pooled non-verbal embeddings as a complement to linguistic features for scalable, privacy-conscious outcome measurement, and provide evidence-based guidance for model selection in small-$n$ affective computing, pending validation in larger routine-care samples. Feature extraction builds on publicly available DISCOVER modules within the NOVA ecosystem, and the full training and model-selection code is released.
The code enables researchers to compute the same feature representation from their own recordings and to rerun the full model-selection procedure on their own labeled data. Reproducing the reported correlations requires a comparably labeled session corpus; neither the recordings nor the trained models can be shared, owing to the sensitive nature of real psychotherapy sessions and the consent given by participants.

\section*{Acknowledgments}
This work was supported by a scholarship of the German Academic Exchange Service (DAAD),
and is partially funded by the Deutsche Forschungsgemeinschaft (DFG, German Research Foundation) -- Grant number 493169211.
We used generative AI (Claude, Codex) for assistance in coding and writing. No labels, data, or experimental results were AI-generated. All decisions and responsibilities rest with the authors.

\bibliographystyle{IEEEtranN}
\bibliography{main}

\end{document}